%% file: main.tex
\newif\ifabstract
\abstracttrue
 \abstractfalse 
\newif\iffull
\ifabstract \fullfalse \else \fulltrue \fi

\documentclass[11pt]{article}

\advance \oddsidemargin by -0.8in
\newcommand{\myparskip}{3pt}
\parskip \myparskip

\usepackage{amsfonts}
\usepackage{amssymb}
\usepackage{amstext}
\usepackage{amsmath}
\usepackage{xspace}
\usepackage{graphicx}
\usepackage{url}
\usepackage{graphics}
\usepackage{colordvi}
\usepackage{colordvi}
\usepackage{subfigure}
\usepackage[utf8]{inputenc}
\usepackage{mathrsfs}
\usepackage{amsmath,amsthm,bm}
\usepackage{xspace}
\usepackage{latexsym}
\usepackage{graphicx}
\usepackage{epsfig}
\usepackage{subfigure}
\usepackage{verbatim}
\usepackage{cases}
\usepackage{url}
\usepackage{amsfonts,latexsym}
\usepackage{setspace}
\usepackage{subfigure}
\usepackage{tikz}
\usepackage{pgfplots}
\usepackage{footnote}
\usepackage{enumitem}
\usepackage{wrapfig}
\usepackage[title]{appendix}

\usepackage{algorithm}
\usepackage[]{algpseudocode}
\makeatletter
\newcommand{\algmargin}{\the\ALG@thistlm}

\makeatother
\newlength{\whilewidth}
\algdef{SE}[parWHILE]{parWhile}{EndparWhile}[1]
{\parbox[t]{\dimexpr\linewidth-\algmargin}{%
		\hangindent\whilewidth\strut\algorithmicwhile\ #1\ \algorithmicdo\strut}}{\algorithmicend\ \algorithmicwhile}%
\algnewcommand{\parState}[1]{\State%
	\parbox[t]{\dimexpr\linewidth-\algmargin}{\strut #1\strut}}

\makeatletter

\newcommand{\Rmnum}[1]{\expandafter\@slowromancap\romannumeral #1@}
\makeatother

\usepackage{array}
\usepackage{color}

\newtheorem{thm}{Theorem}[section]
\newtheorem{lem}[thm]{Lemma}
\newtheorem{pro}[thm]{Proposition}

\newtheorem {dfn}[thm]{Definition}

\newcommand{\cali}{\mathcal{I}}

\newcommand{\calx}{\mathcal{X}}

\newcommand{\calp}{\mathcal{P}}

\newcommand{\la}{{\lambda}}

\newcommand{\opt}{\texttt{OPT}\xspace}
\newcommand{\alg}{\texttt{ALG}\xspace}
\newcommand{\ota}{\texttt{OTA}\xspace}

\newcommand{\comr}{\texttt{CR}\xspace}
\newcommand{\loa}{\texttt{LOA}\xspace}

\newcommand{\otpalgwc}{\texttt{Alg}$(\lambda^{\texttt{w}})$\xspace}
\newcommand{\otpalgoff}{\texttt{Alg}$(\lambda^{\texttt{off}})$\xspace}
\newcommand{\otpalgalf}{\texttt{Alg}$(\lambda^{\texttt{alf}})$\xspace}
\newcommand{\otpalgstat}{\texttt{Alg}$(\lambda^{\texttt{stc}})$\xspace}

\DeclareMathOperator*{\argmax}{arg\,max}

\usepackage{hyperref}
\definecolor{DarkGreen}{RGB}{0, 100, 0}

\usepackage[utf8]{inputenc} 
\usepackage[T1]{fontenc}    
\usepackage{hyperref}       
\usepackage{url}            
\usepackage{booktabs}       
\usepackage{amsfonts}       
\usepackage{nicefrac}       
\usepackage{microtype}      
\usepackage{xcolor}         

\begin{document}

\title{Pareto-Optimal Learning-Augmented Algorithms for Online Conversion Problems}

\author{Bo~Sun\thanks{The Hong Kong University of Science and Technology. Email: {\tt bsunaa@connect.ust.hk}.}\and 
Russell~Lee\thanks{University of Massachusetts Amherst. Email: {\tt rclee@cs.umass.edu}.} \and
Mohammad~Hajiesmaili\thanks{University of Massachusetts Amherst. Email: {\tt hajiesmaili@cs.umass.edu}.} \and
Adam~Wierman\thanks{California Institute of Technology. Email: {\tt adamw@caltech.edu}.} \and
Danny~H.K.~Tsang\thanks{The Hong Kong University of Science and Technology. Email: {\tt eetsang@ust.hk}.}
}

\begin{titlepage}
\maketitle

\thispagestyle{empty}

\begin{abstract}
This paper leverages machine-learned predictions to design competitive algorithms for online conversion problems with the goal of improving the competitive ratio when predictions are accurate (i.e., consistency), while also guaranteeing a worst-case competitive ratio regardless of the prediction quality (i.e., robustness).
We unify the algorithmic design of both integral and fractional conversion problems, which are also known as the 1-max-search and one-way trading problems, into a class of online threshold-based algorithms (OTA). 
By incorporating predictions into design of OTA, we achieve the Pareto-optimal trade-off of consistency and robustness, i.e., no online algorithm can achieve a better consistency guarantee given for a robustness guarantee. 
We demonstrate the performance of OTA using numerical experiments on Bitcoin conversion. 
\end{abstract}

\end{titlepage}

\section{Introductions}
An online conversion problem aims to convert one asset to another through a sequence of exchanges at varying rates in order to maximize the terminal wealth in financial markets. 
With limited information on possible future rates, the core challenge in an online conversion problem is how to balance the return from waiting for possible high rates with the risk that high rates never show up. A high profile example of this risk is cryptocurrency markets, e.g., Bitcoin, where high fluctuations up and down make it challenging to optimize exchanges.
Two well-known classical online conversion problems are 1-max-search~\cite{el1998competitive} and one-way trading~\cite{el2001optimal}, which can be considered as integral and fractional versions of the online conversion problem that trade the asset as a whole or fraction-by-fraction (e.g., trading stock in lot or shares). Beyond these two problems, a number of extensions and variants of online conversion problems have been studied with applications to lookback options~\cite{lorenz2009optimal}, online portfolio selection~\cite{li2014online}, online bidding~\cite{blum2006online}, and beyond.

Most typically, conversion problems are studied through the lens of competitive ratios and the goal is to design online algorithms that minimize the worst-case return ratio of the offline optimal to online algorithm decisions. For example, EI-Yaniv et al.~\cite{el2001optimal} have shown that optimal online algorithms can be designed to achieve the minimal competitive ratios for both 1-max-search and one-way trading. 
However, in real-world problems, predictions about future conversion rates are increasingly available and the algorithms developed in the literature are not designed to take advantage of such information.  The challenge for using such predictions is that, 
in one extreme, the additional information is an accurate prediction (advice) of future inputs.  In this case, the algorithm can confidently use the information to improve performance, e.g., \cite{emek2011online}.
However, most commonly, predictions have no guarantees on their accuracy, and if an online algorithm relies on an inaccurate prediction the performance can be even worse than if it had ignored the prediction entirely.

This challenge is driving the emerging area of the learning-augmented online algorithm (\loa) design, which seeks to design online algorithms that can incorporate untrusted machine-learned predictions in a way that leads to near-optimal performance when predictions are accurate but maintains robust performance when predictions are inaccurate. 
To measure this trade-off, two metrics have emerged, introduced by \cite{lykouris2018competitive} and~\cite{purohit2018improving}: \emph{consistency} and \emph{robustness}.
Consistency is defined as the competitive ratio when the prediction is accurate, i.e., $\comr(0)$, where $\comr(\epsilon)$ is the competitive ratio when the prediction error is $\epsilon$. In contrast, \textit{robustness} is the worst competitive ratio over any prediction errors, i.e., $\max_{\epsilon}\comr(\epsilon)$. Thus, consistency and robustness provide a way to quantify the ability of an algorithm to exploit accurate predictions while ensuring robustness to poor predictions. 

In recent years, a stream of literature has sought to design  robust and consistent \loa for a variety of online problems, such as online caching~\cite{lykouris2018competitive}, ski-rental~\cite{angelopoulos2019online,purohit2018improving,wei2020optimal,bamas2020primal}, online set cover~\cite{bamas2020primal}, secretary and online matching~\cite{antoniadis2020secretary}, metrical task systems~\cite{antoniadis2020online}, and others.
The ultimate goal is to develop algorithms that are \textit{Pareto-optimal} across robustness and consistency, in the sense that for any $\gamma$, the \loa achieves the minimal consistency guarantee among all online algorithms that are $\gamma$-competitive. For the ski rental problem, recent works have derived Pareto-optimal algorithms, e.g., \cite{angelopoulos2019online,wei2020optimal,bamas2020primal}, but in most cases the question of whether there exists a Pareto-optimal \loa is yet to be answered. 

In this paper, we focus on the design of \loa for online conversion problems and we seek to answer the following question:  \emph{Is there a Pareto-optimal \emph{\loa} for the online conversion problem?}

\paragraph{Contributions.} We show that the answer to the above question is ``yes'', by designing an online threshold-based algorithm (\ota), and proving that it is Pareto-optimal.  In particular, we introduce a class of \ota that unifies the algorithmic design of both 1-max-search and one-way trading. We then incorporate predictions into \ota by parameterizing the threshold functions based on the predictions.  This approach yields bounded consistency and robustness (see Theorem~\ref{thm:1-max-search} and Theorem~\ref{thm:one-way-trading}).  Further, we derive lower bounds for robustness-consistency trade-offs and show that our learning-augmented \ota achieves those lower bounds, and is thus Pareto-optimal (see Theorem~\ref{thm:lower-bound-1-max} and Theorem~\ref{thm:lower-bound-otp}).
Finally, we demonstrate the improvement of the learning-augmented \ota over pure online algorithms using numerical experiments based on real-world data tracking Bitcoin prices.

The technical contributions of this paper are twofold. First, 
we provide a sufficient condition for design and analysis of the learning-augmented \ota with a guaranteed generalized competitive ratio. 
This competitive ratio is general in the sense that it not only can yield robustness and consistency guarantees, but can also potentially provide more fine-grained performance guarantees beyond robustness and consistency.
Second, we provide a novel way of deriving the lower bound on the robustness-consistency trade-off, which may be of use beyond online conversion problems. The key idea is to construct a function that can model all online algorithms under a special family of instances, and the lower bound can be derived from combining the robustness and consistency requirements on this function. This constructive approach to arriving at a lower bound is distinctive.

\section{Problem statement and a unified algorithm}
\label{sec:problem-statement}

\paragraph{The online conversion problem.} 
An online conversion problem considers how to convert one asset (e.g., dollars) to another (e.g., yens) over a trading period $[N]:= \{1,\dots,N\}$.
At the beginning of step ${n\in[N]}$, an exchange rate (or price), $v_n$, is announced and a decision maker must immediately determine the amount of dollars, $x_n$, to convert and obtains $v_n x_n$ yens.
The conversion is unidirectional, i.e., yens are not allowed to convert back to dollars.
The trading horizon $N$ is unknown to the decision maker, and if there are any remaining dollars after $N-1$ trading steps, all of them will be compulsorily converted to yens at the last price $v_N$. 
Without loss of generality, the initial asset can be assumed to be $1$ dollar, and the goal is to maximize the amount of yens acquired at the end of the trading period. 
The offline version of the conversion problem can be cast as
\begin{align}\label{p:online-conversion}
	\underset{x_n}{\rm maximize} \quad \sum\nolimits_{n\in[N]} v_n x_n, \quad {\rm subject\ to} \quad \sum\nolimits_{n\in[N]} x_n \le 1.
\end{align}
If the conversion is only allowed in a single transaction, the decision $x_n \in \{0,1\}$ is a binary variable, and this integral version is called \textit{$1$-max-search}~\cite{el1998competitive}.
If the asset is allowed to convert fraction-by-fraction over multiple transactions, the decision $x_n \in [0,1]$ is a continuous variable, and this fractional version is refereed to as \textit{one-way trading}~\cite{el2001optimal}. Following the literature, we assume the prices $\{v_n\}_{n\in[N]}$ are bounded, i.e., $v_n\in[L,U], \forall n\in[N]$, where $L$ and $U$ are known parameters, and define $\theta = U/L$ as the price fluctuation. 


\paragraph{Online threshold-based algorithms.}
Online threshold-based algorithms (\ota) belong to a class of \textit{reserve-and-greedy} algorithms  where the idea is to use a threshold function to determine the amount of resources that need to be \textit{reserved} based on resource utilization, and then \textit{greedily} allocate resources respecting the reservation in each step.


\ota is known to be easy-to-use but hard-to-design due to the difficulties in developing the threshold function. Prior work using \ota has often been problem specific. For example, an optimal design of the threshold function in \ota is derived in~\cite{sun2020competitive} for one-way trading and in~\cite{zhang2017optimal,zhou2008budget}  for the online knapsack problem, which is closely related to one-way trading.

Here, we unify online algorithms for online conversion problems in an \ota framework in Algorithm~\ref{alg:ota-overall}.  Algorithm~\ref{alg:ota-overall}
takes a threshold function $\phi$ as its input, where $\phi(w):[0,1]\to[L,U]$ is a function of resource utilization (i.e., amount of traded dollar) $w$ and $\phi(w)$ can be considered as the reservation price when the utilization is $w$.
The algorithm makes conversions only if the current price $v_n$ is at least $\phi(w^{(n-1)})$, where $w^{(n-1)} = \sum_{i\in[n-1]} \bar{x}_i$ is the utilization after the previous $n-1$ steps of trading. 
More specifically, the conversion decision $\bar{x}_n$ in each step is determined by solving an optimization problem in Line 3 of Algorithm~\ref{alg:ota-overall}. The \ota framework transforms the algorithmic design task in online conversion problems into the design of $\phi$, and the challenge is to design $\phi$ such that \ota can have theoretical performance guarantees.
To provide two examples, in the following we show how to recover the optimal online algorithms for 1-max-search and one-way trading. 

\textit{1-max-search.}
In this integral conversion problem, \ota sets the feasible space as $\calx_n = \{0,1\}$ and the threshold function as a constant $\phi(w) = \Phi, w\in[0,1]$, where $\Phi$ is also called a reservation price. Then the algorithm simply selects the first price that is at least $\Phi$.
When the reservation price is designed as $\Phi = \sqrt{LU}$, $\ota$ is exactly the same algorithm as the reservation price policy in \cite{el2001optimal}, which achieves the optimal competitive ratio $\sqrt{\theta}$. 

\textit{One-way trading.}
\ota sets $\calx_n = [0,1-w^{(n-1)}]$ and $\phi$ as a continuous and strictly increasing function. The conversion decisions fall into three cases based on the solution of the optimization in Line~3: (i) if $v_n < \phi(w^{(n-1)})$, make no conversions, i.e., $\bar{x}_n = 0$; (ii) if $\phi(w^{(n-1)})\le v_n\le \phi(1)$, $\bar{x}_n$ can be solved based on the first-order optimality condition, i.e., $v_n = \phi(w^{(n-1)} + \bar{x}_n)$; and (iii) if $v_n > \phi(1)$, $\bar{x}_n = 1- w^{(n-1)}$ converts all its remaining dollar at the price $v_n$.
By setting the threshold function to $\phi(w) = L + (\alpha^*L - L)\exp(\alpha^* w), w\in[0, 1]$, $\ota$ achieves the optimal competitive ratio $\alpha^* = 1 + W((\theta-1)/e)$,
where $W(\cdot)$ is the Lambert-W function~\cite{sun2020competitive}. 

\begin{algorithm}[t]
	\caption{Online threshold-based algorithm with threshold function $\phi$ ($\ota_\phi$)}
	\label{alg:ota-overall}
	\begin{algorithmic}[1]
		\State \textbf{input:} threshold function $\phi(\cdot)$, and initial resource utilization (i.e., traded dollar) $w^{(0)} = 0$;
		\While{price $v_n$ is revealed}
		\State determine resource allocation $\bar{x}_n= \argmax_{ x_n \in \calx_n} v_n x_n - \int_{w^{(n-1)}}^{w^{(n-1)}+x_n} \phi(u)du$;
		\State update the utilization $w^{(n)} = w^{(n-1)} + \bar{x}_n$.
		\EndWhile
	\end{algorithmic}
\end{algorithm}


\section{Robustness and consistency}
\label{sec:algorithm-design}

This paper is focused on the design of learning-augmented online algorithms (\loa) where the online algorithm is given a machine-learned prediction $P\in[L,U]$ of the maximum price $V = \max_{n\in[N]} v_n$ over the price sequence. The prediction is not necessarily accurate and we define $\epsilon = |V-P|$ as the prediction error. Let $\comr(\epsilon)$ denote the competitive ratio of \ota when the prediction error is $\epsilon$. Our goal is to design an algorithm that is $\eta$-consistent and $\gamma$-robust, i.e., an algorithm where $\eta \ge \comr(0)$ and $\gamma \ge \max_\epsilon \comr(\epsilon)$. We first focus on designing a learning-augmented \ota by incorporating predictions into the design of the threshold function $\phi$ to achieve bounded robustness and consistency.

\subsection{Warmup}
\label{subsec:algorithmic-challenge}

To highlight the challenges of algorithm design in this setting, we start by showing that an intuitive use of predictions can result in poor robustness-consistency guarantees.  Thus, it is of essential importance to take advantage of the problem structure in designing the learning-augmented \ota.

To illustrate this, we consider the design of the reservation price $\Phi_P$ for 1-max-search as an example. 
If we blindly use the prediction of the maximum price by setting $\Phi_P = P$, $\ota$ is indeed offline optimal when the prediction is accurate, and thus $1$-consistent. However, its robustness is the worst possible competitive ratio $\theta$, which is achieved when the prediction is $P = U$ and the actual maximum price is $V = U-\epsilon$, where $\epsilon \to 0$.  In fact, the robustness guarantee approaches $\theta$ with an arbitrarily small prediction error.   

Another intuitive design is to set the reservation price as a linear combination of $P$ and the optimal reservation price for pure online algorithms $\sqrt{LU}$, i.e., $\Phi_P = \la \sqrt{LU} + (1-\la) P$, where $\la\in[0,1]$ is called the \textit{robustness parameter}, indicating the distrust in the prediction. The robustness and consistency of this algorithm is characterized by the following result. 

\begin{pro}\label{lem:1-max-search-example}
	Given $\la\in(0,1]$, \emph{\ota} with the reservation price $\Phi_P = \la \sqrt{LU} + (1-\la) P$ for 1-max-search is $(\la\sqrt{\theta} + (1-\la)\theta)$-robust and $\sqrt{\theta}$-consistent.
\end{pro}

This result highlights that, while the robustness is a linear combination of the optimal competitive ratio $\sqrt{\theta}$ and $\theta$,
the consistency is $\sqrt{\theta}$, which yields no improvement over the optimal competitive ratio except for a special case when $\la = 0$. 


\subsection{A sufficient condition}
\label{sec:sufficient-condition}

Together, the two examples in the previous section highlight some of the challenges associated with balancing robustness and consistency in \ota. Given the challenges, we now focus on developing a general approach for the design and analysis of robust and consistent learning-augmented \ota. 
To do so, we first generalize the competitive ratio from a scalar to a vector, where each element corresponds to a competitive ratio over a subset of instances (see Definition~\ref{dfn:generalized-cr}). 
Since consistency and robustness can be considered as the competitive ratios over the subsets of the predicted instances and the other instances, the competitiveness of \ota can be transformed to robustness-consistency guarantees (see Lemma~\ref{lem:connection}).
This transformation leads us to characterize a general sufficient condition (see Theorem~\ref{thm:ota-sufficient-condition}) on the threshold function of \ota that guarantees a generalized competitive ratio over a given subsets of instances. 
Then, combining Theorem~\ref{thm:ota-sufficient-condition} and Lemma~\ref{lem:connection} gives a general approach for analyzing the consistency and robustness of \ota, which we leverage in the analysis of 1-max-search and one-way trading in the following sections in order to illustrate its applicability. 

To begin, let $\opt(\cali)$ and $\alg(\cali)$ denote the returns of offline optimal and an online algorithm under instance $\cali$, respectively. 
Let $\calp:=\{\calp_1,\dots,\calp_I\}$ be a partition of the set $\Omega$ of all instances. 

\begin{dfn}[Generalized competitive ratio]
	\label{dfn:generalized-cr}
	$\boldsymbol{\alpha}:=(\alpha_1,\dots,\alpha_I)$ is a generalized competitive ratio over $\calp$ if $\alpha_i = \max_{\cali \in\calp_i}\emph{\opt}(\cali)/\emph{\alg}(\cali)$ is the worst-case ratio over $\calp_i$ for all $i\in[I]$. 
\end{dfn}
In online conversion problems, let $\Omega_p \subseteq \Omega$ be a subset, in which each instance has a maximum price $p$. Thus, if we have a prediction $P$ on the maximum price, it means the instance is predicted to belong to $\Omega_P$.
We can show the consistency and robustness of a learning-augmented \ota by proving its generalized competitive ratio over a partition. 
In particular, given prediction $P$, \ota is $\eta$-consistent and $\gamma$-robust if there exists a partition such that $\ota$ is $\eta$-competitive over the subset that contains $\Omega_P$ and $\gamma$-competitive for the remaining subsets. Formally we have the following claim.
\begin{lem}
	\label{lem:connection}
	Given a prediction $P\in[L,U]$, and parameters $\eta$ and $\gamma$ with $\eta\le\gamma$, \emph{\ota} for online conversion problems is $\eta$-consistent and $\gamma$-robust if there exists a partition $\calp = \{\calp_\eta,\calp_\gamma\}$ with $\Omega = \calp_\eta \cup \calp_\gamma$ and $\Omega_P \subseteq \calp_\eta$, and \emph{\ota} is $(\eta,\gamma)$-competitive over $\calp$. 
\end{lem}

Building on Lemma~\ref{lem:connection}, we now focus on how to design the threshold function $\phi$ in \ota to ensure a small generalized competitive ratio. To this end, divide the range of price $[L,U]$ into $I$ price segments $[M_0,M_1),\dots,[M_{I-1},M_I]$ with $L = M_0 < M_1 <...< M_I = U$. 
We partition $\Omega$ based on the price segments, i.e., $\Omega = \{\Omega_p\}_{p\in[M_0,M_1)} \cup \dots \cup \{\Omega_p\}_{p\in[M_{I-1},M_I]}$. Hereafter, we use $\calp_i$ or $[M_{i-1},M_i)$ to denote the $i$-th instance subset $\{\Omega_p\}_{p\in[M_{i-1},M_i)}$.
To ensure different worst-case ratios over different subsets of instances, we consider a piece-wise threshold function $\phi$ created by concatenating a sequence of functions $\{\phi_i\}_{i\in[I]}$, where each piece $\phi_i$ is designed to guarantee $\alpha_i$-competitiveness over $\calp_i$.
In particular, divide the feasible region $[0,1]$ into $I$ resource segments $[\beta_0,\beta_1),\dots,[\beta_{I-1},\beta_I]$ with $0 = \beta_0 \le \beta_1 \le \dots \le \beta_I = 1$, and $\phi_i(w) \in [M_{i-1},M_i), w \in[\beta_{i-1},\beta_i)$.
We say $\phi_i$ is absorbed if $\beta_{i-1} = \beta_i$.
The following theorem then provides a sufficient condition for designing the threshold function $\phi$ in \ota to guarantee a generalized competitive ratio.

\begin{thm}\label{thm:ota-sufficient-condition}
	\emph{\ota} is $\boldsymbol{\alpha}$-competitive over $\{\calp_i\}_{i\in[I]}$ for online conversion problems if $\phi:=\{\phi_i\}_{i\in[I]}$ is a piece-wise and right-continuous function, $\phi(1)\in\{M_i\}_{i\in[I]}$ is one of the partition boundaries, and each threshold piece $\phi_i(w):[\beta_{i-1},\beta_i)\to [M_{i-1},M_i)$ satisfies one of the following conditions: 
	
	\emph{Case I}: if $M_i \le \phi(0)$, then $M_i \le \alpha_i L$ and $\beta_i = 0$;
	
	\emph{Case II}: if $\phi(0) < M_i \le \phi(1)$, then $\phi_i$ is in the form of 
	\begin{align}\label{eq:sufficient-condition-phi}
		\phi_i(w)=
		\begin{cases}
			M_{i-1} & w\in[\beta_{i-1},\beta_{i-1}')\\
			\varphi_i(w) & w\in[\beta_{i-1}',\beta_{i})
		\end{cases},
	\end{align}
	which consists of a flat segment in $[\beta_{i-1},\beta_{i-1}')$ and a strictly increasing segment $\varphi_i(w)$ that satisfies 
	\begin{align}
		\label{eq:sufficient-condition-DE}
		\begin{cases}
			\varphi_i(w) \le \alpha_i\left[\int_{0}^{\beta_{i-1}'} \phi(u)du +\int_{\beta_{i-1}'}^w \varphi_i(u)du+ (1 - w) L\right], \forall w\in[\beta_{i-1}',\beta_i)\\
			\varphi_i(\beta_i) = M_i
		\end{cases};
	\end{align}
	
	\emph{Case III}: if $M_i > \phi(1)$, then $M_i \le \alpha_i\int_{0}^1\phi(u)du$ and $\beta_i = 1$.
\end{thm}

Theorem~\ref{thm:ota-sufficient-condition} is the key to the analysis that follows. In particular, it provides a sufficient condition for analyzing the generalized competitive ratio of \ota, which in turn yields bounds on  
consistency and robustness as highlighted in Lemma~\ref{lem:connection}. We show its broad applicability in the subsections that follow by applying it in the context of 1-max search and one-way trading. 

Further, this approach is general and provides opportunities to derive more fine-grained performance metrics for the learning-augmented \ota beyond consistency and robustness. For example, instead of just focusing on the improved competitive ratio when predictions are accurate, we can redefine the consistency as a prediction-error dependent metric $\kappa(\xi) := \max_{\epsilon\le\xi}\comr(\epsilon)$.
$\kappa(\xi)$ characterizes the improved competitive ratio if the actual value is within the neighbor of the prediction $[P-\xi,P+\xi]$. Thus, $\kappa(\xi)$ is a more general and fine-grained metric, and $\eta$ and $\gamma$ are two extreme points of $\kappa(\xi)$, i.e., $\eta = \kappa(0)$ and $\gamma = \kappa(\infty)$.
Given $\xi$, we can leverage the competitive ratio to design $\kappa(\xi)$-consistent and $\gamma$-robust \ota. 
In particular, given $P$, \ota is $\kappa(\xi)$-consistent and $\gamma$-robust if there exists a partition $\calp = \{\calp_{\kappa(\xi)},\calp_{\gamma}\}$ with $\Omega_{p\in[P-\xi,P+\xi]} \subseteq\calp_{\kappa(\xi)}$, and \ota is 
$(\kappa(\xi),\gamma)$-competitive over $\calp$.

\subsection{1-max search}
\label{subsec:1-max-search}

We now apply the sufficient condition in Theorem~\ref{thm:ota-sufficient-condition} to the setting of 1-max search. Our goal is to design the reservation price $\Phi_P$ given a prediction $P$.
To do this, we set $\eta:=\eta(\lambda)$ and $\gamma:=\gamma(\lambda)$ as
\begin{align}\label{eq:robust-consistent-1-max}
	\gamma(\la) = [{\sqrt{(1-\la)^2+ 4\la\theta} - (1-\la) }]/{(2\la)},\ \text{and}\ \eta(\la) = {\theta}/{\gamma(\la)},
\end{align}
where $\la\in[0,1]$ is the robustness parameter, and $\eta$ and $\gamma$ are predetermined parameters for designing $\Phi_P$ that represent the consistency and robustness that we target to achieve. In particular, $\eta$ and $\gamma$ are designed as the solution of
\begin{align}\label{eq:eta-gamma-design-1-max}
	\eta(\la) = \theta/\gamma(\la),\ \text{and}\
	\eta(\la) = \lambda \gamma(\la) + 1 - \lambda.
\end{align}
The first equation is the desired trade-off between robustness and consistency, which will be shown to match the lower bound in Section~\ref{sec:lower-bound}, and thus represents a Pareto-optimal trade-off.
The second equation sets $\eta$ as a linear combination of $1$ and $\gamma$. In this way, as $\lambda$ increases from $0$ to $1$
, $\eta$ increases from the best possible ratio $1$ to the optimal competitive ratio $\sqrt{\theta}$, and $\gamma$ decreases from the worst possible ratio $\theta$ to $\sqrt{\theta}$.
Taking $\eta$ and $\gamma$ as inputs, we design the reservation price $\Phi_P$ as follows:
\begin{subequations}\label{eq:threshold-1-max}
	\begin{align}
		\label{eq:reservation-price1}
		&\text{when $P\in[L,L\eta)$},\ \Phi_P = L\eta;\\
		\label{eq:reservation-price2}
		&\text{when $P\in[L\eta,L\gamma)$},\ \Phi_P = \la L \gamma  + (1-\la) P/\eta;\\
		\label{eq:reservation-price3}
		&\text{when $P\in[L\gamma,U]$},\ \Phi_P = L\gamma.
	\end{align}
\end{subequations}

The following theorem provides robustness and consistency bounds for this algorithm. The result follows from the general sufficient condition for the class of \ota in Section~\ref{sec:sufficient-condition}. 
Given each reservation price $\Phi_P$, the key step of analysis is to determine a proper partition of instances and then analyze the competitive ratio over each subset, in which $\Phi_P$ can satisfy the sufficient condition in one of the cases in Theorem~\ref{thm:ota-sufficient-condition}. 
Take $\Phi_P$ in~\eqref{eq:reservation-price1} for an example. We partition $[L,U]$ into $[L,L\eta)$ and $[L\eta,U]$ by letting $M_1 = L\eta$. Since $\phi(0) = \phi(1) = \Phi_P= L\eta$, $\Phi_P$ satisfies Case I and Case III for $[L,L\eta)$ and $[L\eta,U]$, respectively, and the corresponding competitive ratios are $\alpha_1 = \Phi_P/L = \eta$ and $\alpha_2 = U/\Phi_P = \theta/\eta = \gamma$. Thus, \ota is $(\eta,\gamma)$-competitive over $[L,L\eta)$ and $[L\eta,U]$. Additionally, the predicted instance $\Omega_P \subseteq \Omega_{p\in[L,L\eta)}$, and thus \ota is $\eta$-consistent and $\gamma$-robust based on Lemma~\ref{lem:connection}.

\begin{thm}\label{thm:1-max-search}
	Given $\la\in[0,1]$, \emph{\ota} with the reservation price in Equation~\eqref{eq:threshold-1-max} for $1$-max-search is $\gamma(\la)$-robust and $\eta(\la)$-consistent, where $\gamma(\la)$ and $\eta(\la)$ are given in Equation~\eqref{eq:robust-consistent-1-max}.
\end{thm}

\begin{figure}[t]
	\begin{minipage}[b]{.65\textwidth}
		\subfigure{\label{fig:reservation-price}\includegraphics[width=.45\textwidth]{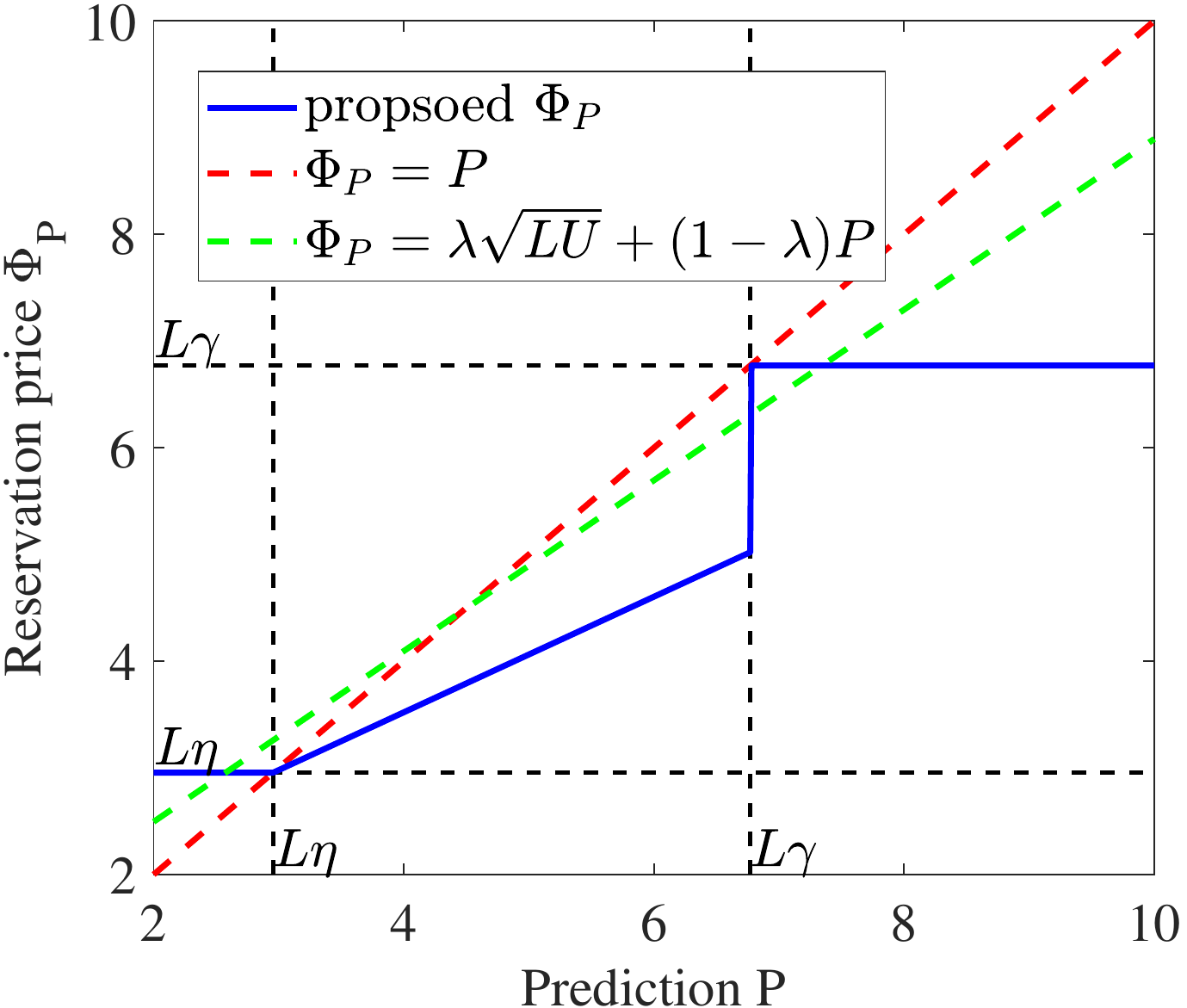}\vspace{0mm}}\hspace{5mm}
		\subfigure{\label{fig:threshold-function-otp}\includegraphics[width=.45\textwidth]{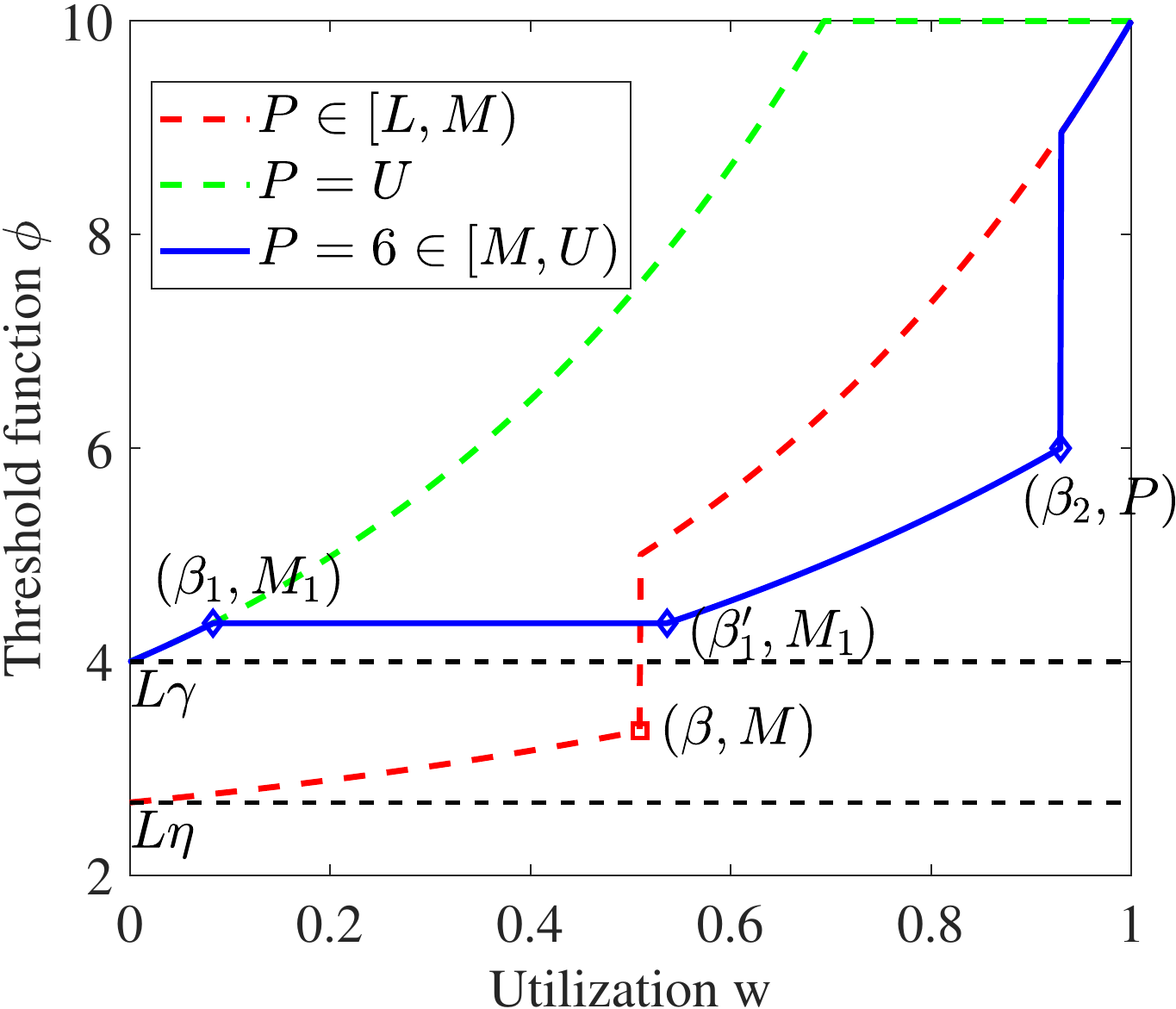}\vspace{0mm}}
		\caption{Reservation price and threshold function for 1-max-search (left) and one-way trading (right) with $L=2$ and $U=10$}
		\label{fig:threshold-function}
	\end{minipage}\hspace{2mm}
	\begin{minipage}[b]{.34\textwidth}\vspace{0mm}
		\includegraphics[width=.97\textwidth]{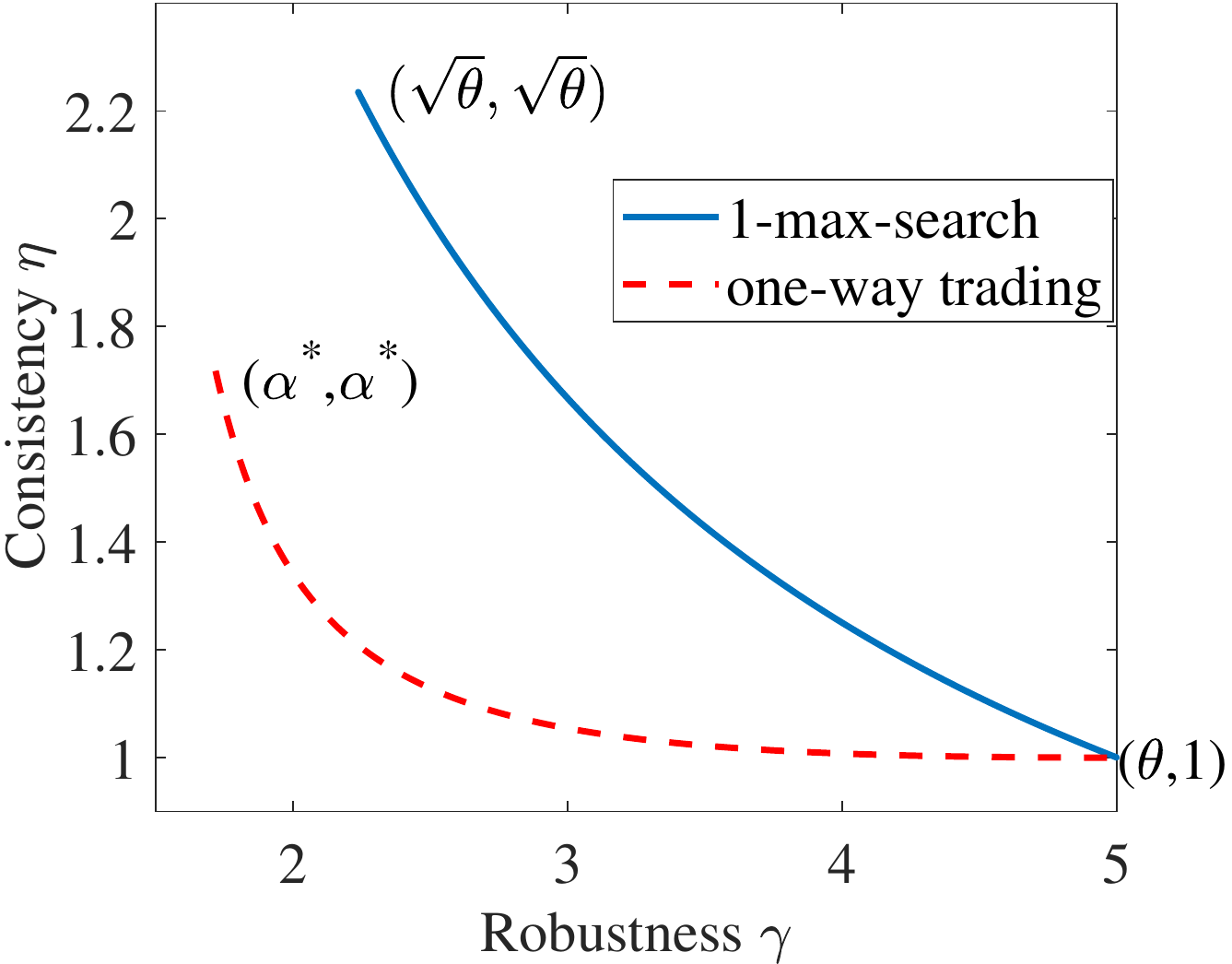}
		\vspace{-1mm}
		\caption{Optimal robustness-consistency trade-offs}
		\label{fig:pareto-boundary}
	\end{minipage}
\end{figure}

Before moving to the proof it is important to give insights into the form of the reservation price~\eqref{eq:threshold-1-max}. It consists of three segments for predictions that are in boundary regions $[L,L\eta)$ and $[L\gamma,U]$ close to price lower and upper bounds, and in intermediate region $[L\eta,L\gamma)$. 
Figure~\ref{fig:threshold-function} illustrates the form and compares it with two intuitive designs $\Phi_P = P$ and $\Phi_P = \la \sqrt{LU} + (1-\la) P$, which we have shown providing poor robustness and consistency guarantees.
Given any reservation price $\Phi\in[L,U]$, the robustness of $\ota$ is $\max\{\Phi/L,U/\Phi\}$, where $\Phi/L$ and $U/\Phi$ are the worst-case ratios over subsets $[L,\Phi)$ and $[\Phi,U]$.
To ensure a good robustness, \eqref{eq:reservation-price1} and~\eqref{eq:reservation-price3} are designed to balance $\Phi/L$ and $U/\Phi$ by just ensuring $\eta$-competitiveness over the boundary region that contains the prediction.
The intuitive design $\Phi_P = P$ neglects this structure, and thus its robustness approaches the worst possible ratio $\theta$.
Given an accurate prediction $P$, the consistency of \ota is $\max\{\Phi/L,P/\Phi\}$.
To guarantee a good consistency, we must avoid the case that $P<\Phi$, leading to the ratio $\Phi/L$ that cannot be properly bounded.
\eqref{eq:reservation-price2} is designed by enforcing $P\ge \Phi_P$. In this way, \ota always makes conversions in the intermediate region and the consistency is $P/\Phi_P$, which can be designed to be upper bounded by $\eta$. 
The intuitive design $\Phi_P = \la \sqrt{LU} + (1-\la) P$ fails to improve the consistency over $\sqrt{\theta}$ since it cannot always guarantee $P\ge \Phi_P$ in the intermediate region, and thus may make no conversions even with an accurate prediction. A full proof of Theorem~\ref{thm:1-max-search} is in Appendix~\ref{app:proof-1-max}.

\subsection{One-way trading}
\label{subsec:one-way-trading}

Next, we apply the sufficient condition in Theorem~\ref{thm:ota-sufficient-condition} to one-way trading. 
We also aim to design the threshold function based on the prediction $P$. 
Here, we set $\gamma:=\gamma(\lambda)$ and $\eta := \eta(\lambda)$ as
\begin{align}\label{eq:robust-consiste-otp}
	\eta(\lambda)= {\theta}/\left[{\frac{\theta}{\gamma(\lambda)}+(\theta - 1)\left(1-\frac{1}{\gamma(\lambda)}\ln\frac{\theta - 1}{\gamma(\lambda) - 1}\right)}\right],\ \text{and}\ \gamma(\la) = \alpha^* + (1-\la)(\theta - \alpha^*),
\end{align}
where $\lambda \in[0,1]$ is the robustness parameter and $\alpha^*$ is the optimal competitive ratio of one-way trading. 
Similarly to the design in 1-max-search, the two equations in~\eqref{eq:robust-consiste-otp} determine the desired trade-off between $\eta$ and $\gamma$, and their desired relationship with $\lambda$.  Again, we derive a lower bound in Section~\ref{sec:lower-bound} showing that this relationship is tight and provides a Pareto-optimal trade-off. 

Taking $\eta$ and $\gamma$ as inputs, we design the threshold function as follows:
\begin{subequations}\label{eq:threshold-otp-P}
	\begin{align}
		\label{eq:threshold-otp-P1}
		&\text{when $P\in[L,M)$},\ \phi_P(w)=
		\begin{cases}
			L + (\eta L - L) \exp(\eta w) & w\in[0,\beta)\\
			L + (U-L)\exp(\gamma(w-1)) &  w\in[\beta,1]
		\end{cases},\\
		\label{eq:threshold-otp-P2}
		&\text{when $P\in[M,U]$},\ \phi_P(w)=
		\begin{cases}
			L + (\gamma L - L) \exp(\gamma w)  & w\in[0,\beta_1)\\
			M_1 &  w\in[\beta_1,\beta_1') \\
			L + (M_1 - L) \exp(\eta(w-\beta_1')) & w\in[\beta_1',\beta_2]\\
			L + (U-L)\exp(\gamma(w-1)) & w\in(\beta_2,1]
		\end{cases},
	\end{align}
\end{subequations}
where $\beta$ and $M$ are solutions of 
\begin{align}\label{eq:otp-threshold-breakpoints1}
	\begin{cases}
		M = L + (\eta L - L) \exp(\eta \beta),\\
		{M}\gamma/{\eta} = {L + (U-L)\exp(\gamma(\beta-1))};
	\end{cases}
\end{align}
and $M_1$, $\beta_1$, $\beta_1'$, and $\beta_2$ are all functions of $P$ and are determined by
\begin{align}\label{eq:otp-threshold-breakpoints2}
	\begin{cases}
		\beta_1 = \frac{1}{\gamma} \ln\frac{\max\{M_1/L,\gamma\}-1}{\gamma - 1},\\
		\frac{M_1}{\eta}=\int_{0}^{\beta_1}\phi(u)du + (\beta_1'-\beta_1)M_1 + (1-\beta_1')L, \\
		P=L + (M_1 - L) \exp(\eta(\beta_2-\beta_1')),\\
		\beta_2= 1 + \frac{1}{\gamma}\ln\frac{\min\{P\gamma/\eta,U\}-L}{U-L}.
	\end{cases}
\end{align}

The following theorem provides robustness and consistency bounds for this algorithm. Again, the result follows from the general sufficient condition for the class of \ota that we introduce in Section~\ref{sec:sufficient-condition}.  
Compared to 1-max-search, the additional difficulty of one-way trading lies in the analysis of the competitive ratios over the subsets belonging to Case II of the sufficient condition since the threshold function $\phi_P$ ranging in these subsets needs to satisfy a set of differential equations~\eqref{eq:sufficient-condition-DE}. 
$\phi_P$ in~\eqref{eq:threshold-otp-P} is in fact designed as the solution of the differential equation~\eqref{eq:sufficient-condition-DE} with binding inequalities and properly designed boundary conditions (by setting the length of the flat segment of each threshold piece).

\begin{thm}\label{thm:one-way-trading}
	Given $\lambda\in[0,1]$, \emph{\ota} with the threshold function~\eqref{eq:threshold-otp-P} for one-way trading is $\gamma(\lambda)$-robust and $\eta(\lambda)$-consistent, where $\gamma(\lambda)$ and $\eta(\lambda)$ are given in Equation~\eqref{eq:robust-consiste-otp}. 
\end{thm}

Figure~\ref{fig:threshold-function} illustrates the function given different predictions. The basic idea behind the design of the threshold function~\eqref{eq:threshold-otp-P} is similar to that of 1-max-search.
When the prediction is in the boundary region $P\in[L,M)$, the threshold function~\eqref{eq:threshold-otp-P1} (i.e., red curve) is designed to ensure $\eta$-competitiveness over $[L,M)$, and additionally guarantee $\gamma$-competitiveness over $[M,U]$. 
In the other extreme when $P = U$, the threshold function~\eqref{eq:threshold-otp-P2} becomes $\phi_P(w) = L + (L\gamma-L)\exp(\gamma w), w\in[0,\beta_1)$ and $\phi_P(w) = U, w\in[\beta_1,1]$ (i.e., green curve) since $\beta_2 = \beta_1' = 1$ and $M_1 = U$ by solving equation~\eqref{eq:otp-threshold-breakpoints2} with $P = U$.
This threshold is $(\eta,\gamma)$-competitive over $[U]$ and $[L,U)$.
When the prediction is in the intermediate region $P\in[M,U)$, the threshold function consists of at most  four segments. The first and the forth segments when $w\in[0,\beta_1)$ and $w\in(\beta_2,1]$ are exponential functions with rate $\gamma$, aiming to ensure $\gamma$-competitiveness over $[L,M_1)$ and $(P,U]$. These two segments may be absorbed when the prediction is small ($M_1\le L\gamma$) or large ($P\gamma/\eta \ge U$), corresponding to $\beta_1 = 0$ and $\beta_2 = 1$.
To guarantee a good consistency, a flat segment in $w\in[\beta_1,\beta_1')$ is designed to convert enough dollar before reaching the price $P$ by enforcing $P\ge M_1$, and an exponential segment with rate $\eta$ in $w\in[\beta_1',\beta_2]$ to ensure $\eta$-competitiveness over $[M_1,P]$. 
A full proof can be found in Appendix~\ref{app:proof-one-way-trading}.

\section{Pareto-optimal consistency-robustness trade-off}
\label{sec:lower-bound}

To this point, we have focused on upper bounds for robustness and consistency.  This section provides lower bounds on the robustness-consistency trade-offs for both 1-max-search and one-way trading and shows the Pareto-optimality of our proposed learning-augmented algorithms.  Note that, obtaining lower bounds on the trade-off between robustness and consistency for online algorithms has proven difficult.  The only existing tight lower bounds we are aware of are in the case of deterministic~\cite{angelopoulos2019online} and randomized~\cite{bamas2020primal,wei2020optimal} algorithms for the ski-rental problem.   

\begin{thm}
	\label{thm:lower-bound-1-max}
	Any $\gamma$-robust deterministic \emph{\loa}
	for 1-max-search must have consistency $\eta \ge {\theta}/{\gamma}$. Thus, \emph{\ota} with the reservation price~\eqref{eq:threshold-1-max} is Pareto-optimal.
\end{thm}

\begin{thm}
	\label{thm:lower-bound-otp}
	If a deterministic \emph{\loa}
	for one-way trading is $\gamma$-robust, its consistency is at least ${\eta \ge {\theta}/[\frac{\theta}{\gamma}+(\theta - 1)(1-\frac{1}{\gamma}\ln\frac{\theta- 1}{\gamma  - 1})]}$. Thus,
	\emph{\ota} with the threshold function~\eqref{eq:threshold-otp-P} is Pareto-optimal.
\end{thm}

We illustrate the Pareto-optimal trade-offs of robustness and consistency for 1-max-search and one-way trading in Figure~\ref{fig:pareto-boundary}. 
Notice that the Pareto-boundary of one-way trading dominates that of 1-max-search since the fractional conversion leaves more flexibility to online decisions in one-way trading, leading to a better lower bound.
For both problems, with the improvement of consistency from the optimal competitive ratio (i.e., $\sqrt{\theta}$ or $\alpha^*$) to the best possible ratio $1$, the robustness degrades from the optimal competitive ratio to the worst possible ratio $\theta$. This means there is no free lunch in online conversion problems; to achieve a good consistency, robustness must be sacrificed. 

We end the section by proving Theorem~\ref{thm:lower-bound-otp} for one-way trading.  A proof of Theorem~\ref{thm:lower-bound-1-max} is included in Appendix
\ref{app:proof-lower-bound-1-max}.



\paragraph{Proof of Theorem~\ref{thm:lower-bound-otp}.}
To show a lower bound result, we first construct a special family of instances, and then show that for any $\gamma$-robust \loa (not necessarily being \ota), their consistency $\eta$ is lower bounded under the special instances.

We focus on a collection of $p$-instances $\{\cali_p\}_{p\in[L,U]}$ where $p$ ranges from $L$ to $U$, where a $p$-instance is defined as follows.
\begin{dfn}[$p$-instance]
	Given $p\in[L,U]$ and a large $N$, an instance $\cali_p :=\{v_1,\dots,v_N\}$ is called a $p$-instance if $v_n = L + (n-1)\delta, n\in[N-1]$ with $\delta = \frac{p-L}{N-2}$ and $v_N = L$.
\end{dfn}
Notice that, when $N\to\infty$, the sequence of prices in $\cali_p$ continuously increases from $L$ to $p$, and drops to $L$ in the last step. 


Let $g(p):[L,U]\to [0,1]$ denote a conversion function of a deterministic \loa for one-way trading, where $g(p)$ is its total amount of converted dollar under the instance $\cali_p$ before the compulsory conversion in the last step.
A key observation is that for a large $N$, executing the instance $\cali_{p+\delta}$ is equivalent to first executing $\cali_{p}$ (excluding the last step) and then processing $p+\delta$ and $L$.  
Since the conversion decision is unidirectional and deterministic, we must have $g(p+\delta) \ge g(p)$, i.e., $g(p)$ is non-decreasing in $[L,U]$. In addition, the whole dollar must be converted once the maximum price $U$ is observed, i.e., $g(U) = 1$.

Under the instance $\cali_p$, the offline optimal profit is $\opt(\cali_p) = p$ and the profit of an online algorithm with conversion function $g$ is $ \alg(\cali_p) = g(L)L + \int_{L}^p u dg(u) + L(1-g(p))$,
where $u dg(u)$ is the profit of converting $d g(u)$ dollar at the price $u$. The first two terms are the cumulative profit before the last step and the last term is from the compulsory conversion.

For any $\gamma$-robust online algorithm, the corresponding conversion function must satisfy $${\alg(\cali_p) \ge \opt(\cali_p)/\gamma =  p/\gamma, \forall p\in[L,U]}.$$ 
If, additionally, given prediction $P \ge \gamma L$, no dollar needs to be converted under instances $\{\cali_p\}_{p\in[L,\gamma L)}$, i.e., $g(p)=0, \forall p\in[L,\gamma L)$.
This is because if a $\gamma$-robust online algorithm converts any dollar below the price $\gamma L$, we can always design a new algorithm by letting it convert the dollar at the price $\gamma L$ instead. The new online algorithm is still $\gamma$-robust and achieves a smaller consistency when the prediction is accurate.
Thus, given $P = U \ge L\gamma$, the conversion function of any $\gamma$-robust online algorithm must satisfy
\begin{align}\label{eq:robustness}
	\alg(\cali_p) = g(\gamma L)\gamma L + \int_{\gamma L}^p u dg(u) + L(1-g(p)) \ge \frac{p}{\gamma}, \quad \forall p\in[\gamma L,U].
\end{align}
By integral by parts and Gronwall's Inequality (see Theorem 1, p.356, \cite{Mitrinovic1991}), a necessary condition for above robustness constraint~\eqref{eq:robustness} to hold is
\begin{align}\label{eq:robustness-necessary}
	g(p)\ge \frac{p/\gamma-L}{p-L} + \frac{1}{\gamma}\int_{\gamma L}^p \frac{u-\gamma L}{(u-L)^2} du =  \frac{1}{\gamma}\ln\frac{p-L}{\gamma L-L}, \quad \forall p\in[\gamma L, U].
\end{align}
In addition, to ensure $\eta$-consistency when the prediction is $P = U$, we must ensure ${\alg(\cali_U) \ge \opt(\cali_U)/\eta}$. Combining this constraint with $g(U)=1$ gives
\begin{align}\label{eq:consistency}
	\int_{\gamma L}^U g(u)du \le (\eta - 1)U/\eta.
\end{align}

By combining equations~\eqref{eq:robustness-necessary} and~\eqref{eq:consistency}, the conversion function $g(p)$ of any $\gamma$-robust and $\eta$-consistent online algorithm given $P = U$ must satisfy 
$\int_{\gamma L}^U \frac{1}{\gamma}\ln\frac{u-L}{\gamma L-L} du \le \int_{\gamma L}^U g(u)du \le (\eta - 1)U/\eta,$
which equivalently gives $\eta \ge {\theta}/[\frac{\theta}{\gamma}+(\theta - 1)(1-\frac{1}{\gamma}\ln\frac{\theta- 1}{\gamma  - 1})]$.

Finally, since Theorem~\ref{thm:one-way-trading} has shown that \ota with the threshold function~\eqref{eq:threshold-otp-P} can achieve the lower bound in Theorem~\ref{thm:lower-bound-otp}, it is Pareto-optimal.





\vspace{-0.2cm}
\section{Numerical results}
We end with a case study on the exchange of Bitcoin (BTC) to USD. This case study is timely since the rapid growth of cryptocurrency has left many traders eager to profit from rising and falling of exchange rates in average-case scenarios, while the uncertainty and volatility of cryptocurrency have made many traders cautious of unforeseeable crashes in worst-case scenarios.  
Our results answer two questions: 
(Q1) \textit{How does the learning-augmented \emph{\ota} compare to pure online algorithms with different prediction qualities and drastic exchange rate crashes?}
(Q2) \textit{How should the \emph{\ota} select the robustness parameter $\la$, and especially, would an online learning algorithm work in practice?}

\paragraph{Setup.} We use historical BTC prices in USD of 5 years from 2015 to  2020, with exchange rates collected every 5 minutes from the Gemini exchange. We assume one BTC is available for trading during 250 instances of length one week. To generate a prediction $P$, we simply use the observed maximum exchange rate of the previous week. To evaluate the impact of prediction quality, we adjust the error level between 0.0 to 1.0, where 0.0 indicates perfect predictions and 1.0 indicates unadjusted predictions. To evaluate the performance in worst-case settings, we also introduce a \textit{crash probability} $q$, where the exchange rate of BTC at the last slot will crash to $L$ with probability~$q$. 

We compare the empirical profit ratio of four different algorithms: (i) \otpalgwc, the worst-case optimized online algorithm that does not take into account predictions, but, guarantees the optimal competitive ratio; (ii) \otpalgoff, an
algorithm that finds the best possible distrust parameter $\la$ in an offline manner; this algorithm is not practical since it is fed with the optimal parameter; however, it illustrates the largest possible improvement from predictions under our algorithm; (iii) \otpalgalf, an online learning algorithm from \cite{maillard2010} which selects the parameter using the adversarial Lipschitz algorithm in a full-information setting; and (iv) \otpalgstat, an online algorithm that uses the best static $\la$ and serves as the baseline for \otpalgalf. Additional details are in the supplementary material.

\begin{figure*}[t]
	\subfigure[]{\label{fig-boxplot}\includegraphics[width=.245\textwidth]{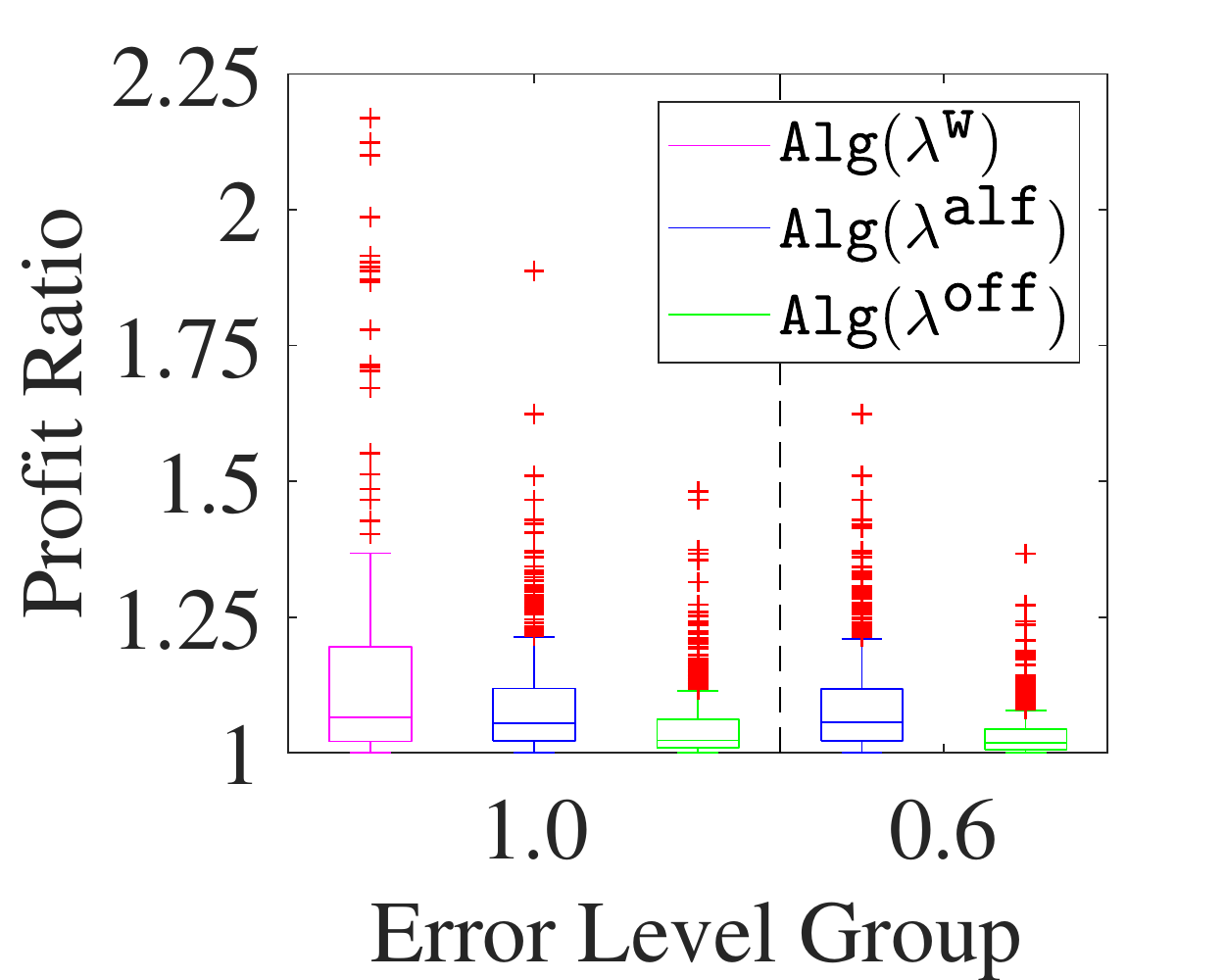}\vspace{0mm}}
	\subfigure[]{\label{fig-boxplot-group}\includegraphics[width=.245\textwidth]{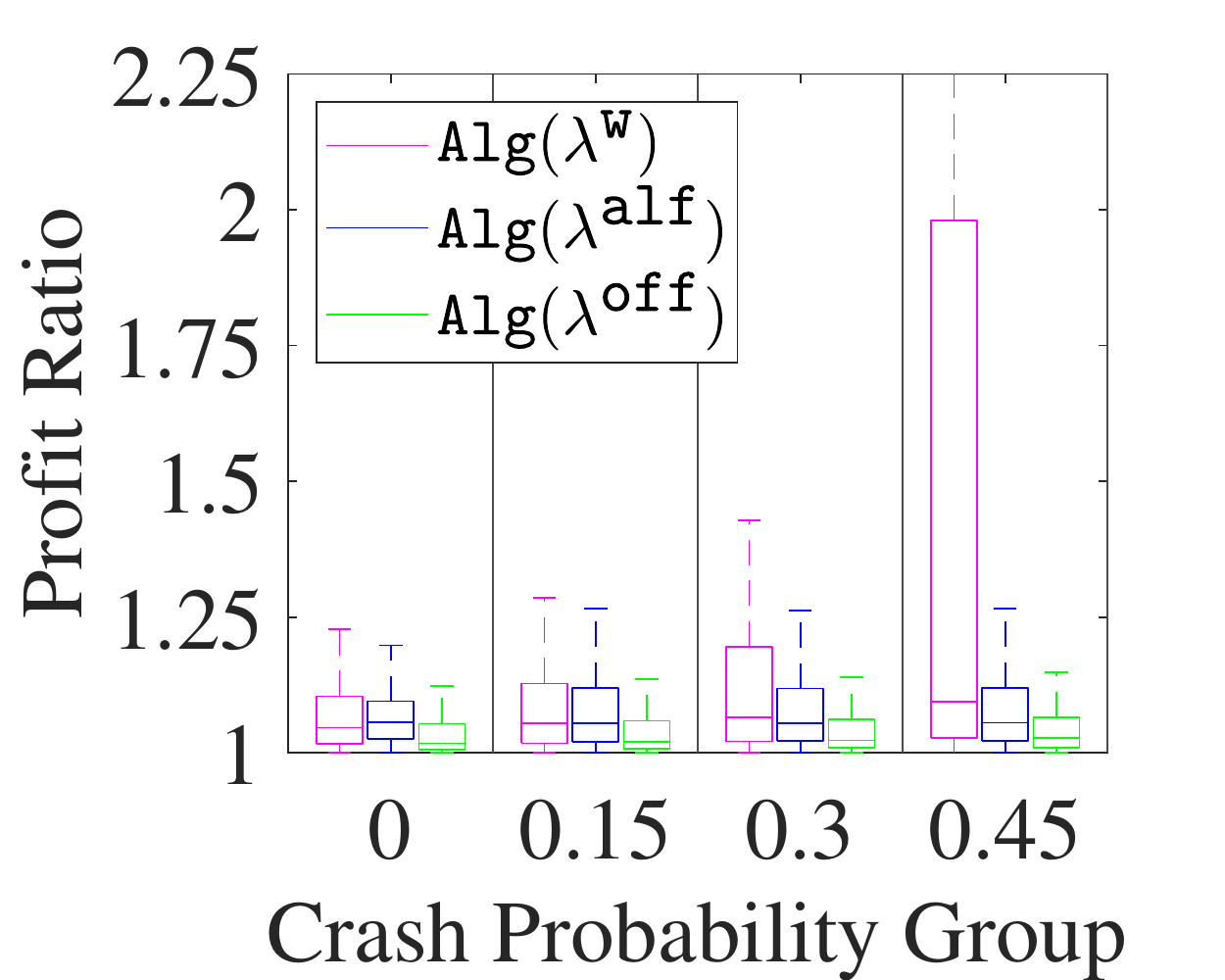}\vspace{0mm}}
	\subfigure[]{\label{fig-reward}\includegraphics[width=.245\textwidth]{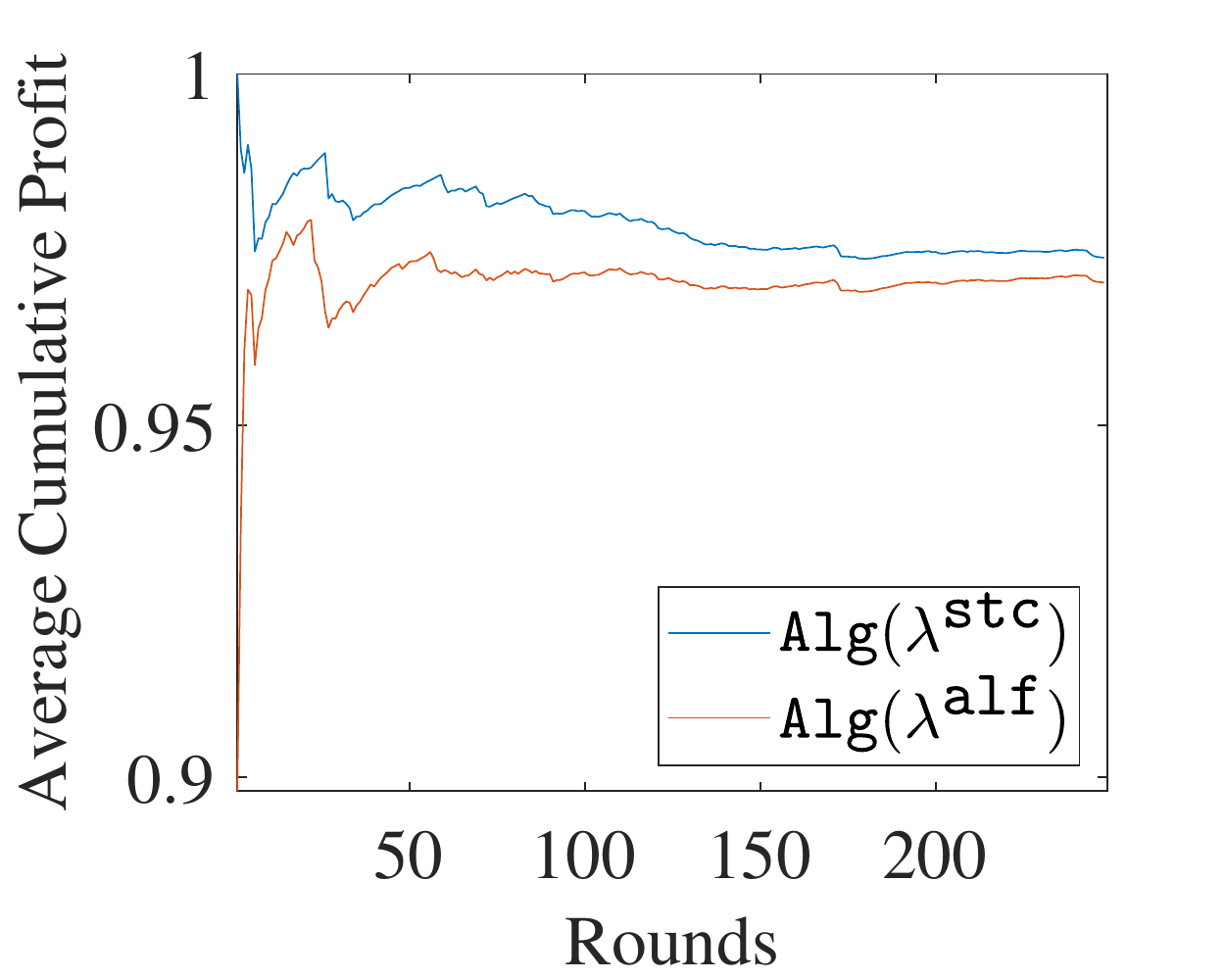}\vspace{0mm}}
	\label{fig:1day_cdfs}
	\subfigure[]{\label{fig-regret}\includegraphics[width=.245\textwidth]{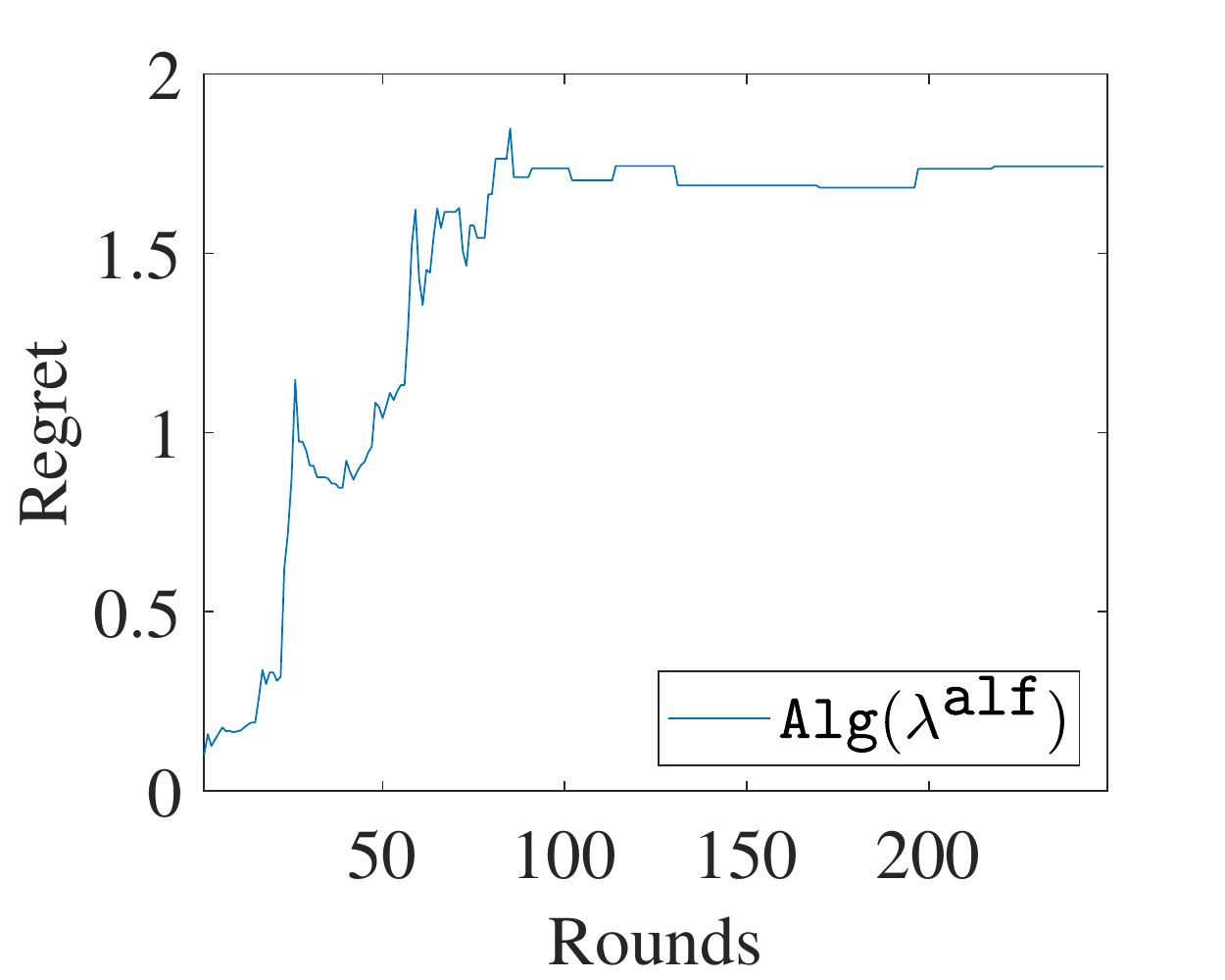}\vspace{0mm}}
	\vspace{-5mm}
	\caption{ Profit ratios of different algorithms with (a) different prediction errors and (b) different crash probabilities. The evolution of (c) the average cumulative profit and (d) regret of \otpalgalf}
	\label{fig:1day_cdfs}
\end{figure*}


\paragraph{Experimental results.} 
We answer Q1 in Figures~\ref{fig-boxplot} and~\ref{fig-boxplot-group}, and Q2 in Figures~\ref{fig-reward} and~\ref{fig-regret}. 
Figure~\ref{fig-boxplot} compare the profit ratios of several algorithms at different error levels.  First, it shows that \otpalgoff and \otpalgalf noticeably improve the performance of \otpalgwc. The upper boxplot whisker of \otpalgwc is 1.35, while \otpalgalf at 1.0 error level has an upper boxplot whisker around 1.25.  Second, it shows that the gap between \otpalgoff and \otpalgalf is quite small, as the upper boxplot whisker of \otpalgoff is slightly lower.  
Comparing the profit ratios of different algorithms with different crash probability values at 1.0 error level in Figure~\ref{fig-boxplot-group}, we see that the performance of \otpalgwc drastically degrades at crash probability 0.45. However, both \otpalgalf and \otpalgoff are stable at high crash probability.
Figure~\ref{fig-reward} compares the average normalized profit of \otpalgalf and \otpalgstat and shows the reward of \otpalgalf converges toward that of \otpalgstat as the learning process moves forward. Figure~\ref{fig-regret} indicates that the regret of \otpalgalf stabilizes.




\section{Concluding remarks}
\label{sec:conclusions}
To improve upon the performance of algorithms for online conversion problems that are designed with worst-case guarantees in mind, this paper has incorporated machined-learned predictions into the design of a class of \ota and shown that the learning-augmented \ota can achieve Pareto-optimal robustness-consistency trade-offs.    
This result represents only the second tight lower bound result in the robustness and consistency analysis of \loa, with the first being for ski rental problem~\cite{angelopoulos2019online,bamas2020primal,wei2020optimal}. 
We expect that our method of deriving lower bounds can be extended to more general online optimization problems with capacity constraints.  
A limitation of this work is that consistency and robustness only measure the competitive performance in two extreme cases, i.e., the predictions are perfectly accurate or completely wrong. Although our approach for design and analysis of \ota provides opportunities to design \ota in ways that guarantee more fine-grained performance metrics, this is left for the future work. 
Another limitation is that this work provides only empirical evaluation of the algorithm \otpalgalf that selects the robustness parameter in an online manner. Deriving theoretical bounds remains open. Last, we cannot see any negative societal impacts of our work.


\bibliographystyle{abbrv} 
\bibliography{reference}
\clearpage

\input{Supplemental_materials}

\end{document}

%% file: Supplemental_materials.tex
\newpage

\begin{appendices}

\section{Technical Proofs}

\subsection{Proof of Proposition~\ref{lem:1-max-search-example}}
\label{app:proof-proposition1}

To see the robustness of \ota with the reservation price $\Phi_P = \la\sqrt{LU}+(1-\la)P$ for 1-max-search, we consider the following two cases.

\textbf{Case I.} When the actual maximum price $V$ is smaller than the reservation price, i.e., $V< \Phi_P$, $1$ dollar is converted at the last step with the worst possible price $L$, and thus the worst-case ratio is 
\begin{align}
    \frac{\opt(\cali)}{\alg(\cali)} = \frac{V}{L} < \frac{\Phi_P}{L} \le \la\sqrt{\theta} + (1-\la)\theta,
\end{align}
where the last inequality is due to $P\le U$.

\textbf{Case II.} When $V \ge \Phi_P$, the worst-case ratio is
\begin{align}
    \frac{\opt(\cali)}{\alg(\cali)} = \frac{V}{\Phi_P} \le \frac{\theta}{\la\sqrt{\theta} + 1-\la},
\end{align}
where the inequality is due to $V\le U$ and $P\ge L$.

Thus, the robustness is $\max\{\la\sqrt{\theta} + (1-\la)\theta,\theta/(\la\sqrt{\theta} + 1-\la)\} = \la\sqrt{\theta} + (1-\la)\theta$.

To see the consistency, consider the following two cases when the prediction is accurate, i.e., $P=V$. 

\textbf{Case I.} When $P \ge \sqrt{LU}$, we have $V = P \ge \Phi_P$ for $\la\in[0,1]$, and the worst-case ratio is
\begin{align}
    \frac{\opt(\cali)}{\alg(\cali)} = \frac{V}{\Phi_P} = \frac{V}{\la\sqrt{LU} + (1-\la)V} \le \frac{\theta}{\la\sqrt{\theta} + (1-\la)\theta}.
\end{align}

\textbf{Case II.} When $P < \sqrt{LU}$, we have $V = P < \Phi_P$ for $\la\in(0,1]$, and the worst-case ratio 
\begin{align}
    \frac{\opt(\cali)}{\alg(\cali)} = \frac{V}{L} < \frac{\la\sqrt{LU} + (1-\la)V}{L}< \sqrt{\theta}.
\end{align}

Combining above two cases gives the consistency $\max\{\sqrt{\theta},\theta/(\la\sqrt{\theta} + (1-\la)\theta)\} = \sqrt{\theta}$ for $\la\in(0,1]$. In the special case when $\la = 0$, $\Phi_P = P = V$ and the consistency is $1$.

\subsection{Proof of Lemma~\ref{lem:connection}}
\label{app:connection}

Recall that given a prediction $P$ on the maximum price $V$ over the price sequence $\cali$, the prediction error is $\epsilon = 0$ if $\cali \in \Omega_P$. Thus, we have $\comr(0) = \max_{\cali\in\Omega_P} \opt(\cali)/\alg(\cali)$. In addition, arbitrary prediction errors mean that the actual instance $\cali$ can take any possible instances in $\Omega$. Therefore, we have $\max_{\epsilon}\comr(\epsilon) = \max_{\cali\in\Omega}\opt(\cali)/\alg(\cali)$.

If \ota is $(\eta,\gamma)$-competitive over a partition $\{\calp_\eta,\calp_\gamma \}$ with $\Omega = \calp_\eta \cup \calp_\gamma$ and  $\Omega_P \subseteq \calp_\eta$, based on the definition of the generalized competitive ratio, we have 
\begin{align}
    \gamma &\ge \max\{\eta,\gamma\} =  \max_{\cali\in \calp_\eta \cup \calp_\gamma} \frac{\opt(\cali)}{\alg(\cali)} = \max_{\epsilon} \comr(\epsilon),\\
    \eta &= \max_{\cali \in \calp_\eta} \frac{\opt(\cali)}{\alg(\cali)} \ge \max_{\cali \in \Omega_P} \frac{\opt(\cali)}{\alg(\cali)} = \comr(0).
\end{align}
Thus, \ota is $\eta$-consistent and $\gamma$-robust.

\subsection{Proof of Theorem~\ref{thm:ota-sufficient-condition}}
\label{app:proof-theorem-sufficient-cond}

Our goal is to prove that $\ota$ with the threshold function ${\phi := \{\phi_i\}_{i\in[I]}}$ is $\alpha_i$-competitive over subset $\calp_i = \{\Omega_p\}_{p\in[M_{i-1},M_i)}$ for all $i\in[I]$ when each piece of the threshold function, $\phi_i$, satisfies the condition in one of the cases of Theorem~\ref{thm:ota-sufficient-condition}. 
Consider the instance subset in the following three cases, corresponding to the three cases of the sufficient condition in Theorem~\ref{thm:ota-sufficient-condition}. 

\textbf{Case I.} $\calp_i$ with $M_i \le \phi(0)$. 
For any instance $\cali \in \calp_i$, the maximum price is smaller than $M_i$ and hence $\opt(\cali) < M_i$. 
Since $\beta_i = 0 = \beta_{i-1}$, this threshold piece is absorbed. Because $\phi$ is a right-continuous function, the reservation price at $\beta_i$ is no smaller than $M_i$.
Thus, \ota converts no dollar when executing $\cali$, excluding the compulsory conversion in the last step, and its return is $\alg(\cali) = v_N \ge L$.
In this case, when the threshold function $\phi_i$ and the competitive ratio $\alpha_i$ satisfy $M_i \le \alpha_i L$,
$\opt(\cali)/\alg(\cali)< M_i/L \le \alpha_i, \forall \cali \in\calp_i$.  

\textbf{Case II.} $\calp_i$ with $\phi(0) < M_i \le \phi(1)$. 
Without loss of generality, we only consider the instances whose price in the last step is not the unique maximum price. 
This is because 
we can instead consider an alternative instance, which appends an additional maximum price just after the unique maximum price as the last price, and this alternative instance leads to the same offline and online returns as the original instance when being executed by \ota. Thus, we can use $w^{(N-1)}$ to denote the final utilization of \ota after executing any instance.

Under an instance $\cali\in\Omega_p\subseteq\calp_i$ with a maximum price $p$, the return of offline optimal is $\opt(\cali) = p$, and the return of \ota is 
\begin{align}
    \alg(\cali) &= \sum_{n\in[N-1]} v_n\bar{x}_n  + (1-w^{(N-1)}) v_N\\
    \label{eq:proof-sufficient-cond1}
    &\ge \sum_{n\in[N-1]}\int_{w^{(n-1)}}^{w^{(n)}} \phi(u)du +(1-w^{(N-1)}) L\\
    &= \int_{0}^{w^{(N-1)}} \phi(u)du +(1-w^{(N-1)})L,
\end{align}
where $\alg(\cali)$ consists of the return of conversions $\{\bar{x}_n\}_{n\in[N-1]}$ by \ota and the compulsory conversion $1-w^{(N-1)}$ in the last step. The inequality~\eqref{eq:proof-sufficient-cond1} holds since (i) $v_N\ge L$, and (ii) $\bar{x}_n$ is the optimal solution of the optimization Line 3 in Algorithm~\ref{alg:ota-overall}, which ensures $v_n \bar{x}_n \ge \int_{w^{(n-1)}}^{w^{(n)}}\phi(u)du$.

In this case, $\phi_i$ is in the form of Equation~\eqref{eq:sufficient-condition-phi}, which consists of a flat segment in $[\beta_{i-1},\beta_{i-1}')$ and an increasing segment $\varphi_i(w)$ in $[\beta_{i-1}',\beta_i)$ that satisfies the differential equation~\eqref{eq:sufficient-condition-DE}.
Note that $w^{(N-1)}$ is the final utilization of \ota after executing $\cali\in\Omega_p$ and $w^{(N-1)}\in[\beta_{i-1}',\beta_i)$.
Also noticing that $\phi$ can be discontinuous at $w = \beta_{i-1}'$, we further consider two sub-cases.

\textbf{Case II(a).} if $M_{i-1}\le p<\varphi_i(\beta_{i-1}')$, then $w^{(N-1)} = \beta_{i-1}'$. 
In this case, we have 
\begin{align*}
 \frac{\opt(\cali)}{\alg(\cali)} \le \frac{p}{\int_{0}^{w^{(N-1)}} \phi(u)du +(1-w^{(N-1)})L} < \frac{\varphi_i(\beta_{i-1}')}{\int_{0}^{\beta_{i-1}'} \phi(u)du +(1-\beta_{i-1}')L} \le \alpha_i,
\end{align*}
where the last inequality is due to the differential equation~\eqref{eq:sufficient-condition-DE} at $w = \beta_{i-1}'$.

\textbf{Case II(b).} if $\varphi_i(\beta_{i-1}')\le p<\varphi_i(\beta_{i}) = M_i$, we have $p = \varphi_i(w^{(N-1)})$ and 
\begin{align*}
 \frac{\opt(\cali)}{\alg(\cali)} \le \frac{p}{\int_{0}^{w^{(N-1)}} \phi(u)du +(1-w^{(N-1)})L} = \frac{\varphi_i(w^{(N-1)})}{\int_{0}^{w^{(N-1)}} \phi(u)du +(1-w^{(N-1)})L} \le \alpha_i, 
\end{align*}
where the last inequality is also due to the differential equation~\eqref{eq:sufficient-condition-DE}.

Combining the above two sub-cases gives $\opt(\cali)/\alg(\cali) \le \alpha_i, \forall \cali\in\Omega_p, \forall \Omega_p\subseteq\calp_i$.

\textbf{Case III.} $\calp_i$ with $M_{i} > \phi(1)$.
Since $\phi(1)$ is one of the price segment boundaries, $M_{i-1}\ge\phi(1)$ and hence $\beta_{i-1} = 1$. Thus, $\phi_i$ is absorbed. 
For any instance $\cali\in\calp_i$, its maximum price is no smaller than the maximum value of the threshold function $\phi(1)$, and thus the whole dollar will be converted to yens before the compulsory conversion, i.e., $w^{(N-1)} = 1$.
The return of the offline optimal is $\opt(\cali) \le M_i$ and the return of \ota is $\alg(\cali) \ge \int_{0}^1 \phi(u)du$. 
In this case, when $\phi$ and $\alpha_i$ satisfy $M_i \le \alpha_i \int_0^1\phi(u)du$, we have $\opt(\cali)/\alg(\cali) \le M_i/\int_{0}^1\phi(u)du \le \alpha_i, \forall \cali\in\calp_i$.

In summary, $\opt(\cali)/\alg(\cali) \le \alpha_i, \forall \cali\in\calp_i$ and thus \ota is $\alpha_i$-competitive over $\calp_i$, $\forall i\in[I]$ if each piece of the threshold function of \ota satisfies one of the sufficient conditions in Theorem~\ref{thm:ota-sufficient-condition}.

\subsection{Proof of Theorem~\ref{thm:1-max-search}}
\label{app:proof-1-max}

We show the competitiveness of $\ota$ with the reservation price~\eqref{eq:threshold-1-max} for 1-max-search based on the sufficient condition in Theorem~\ref{thm:ota-sufficient-condition} and further prove the robustness and consistency bounds based on Lemma~\ref{lem:connection}. 
Consider the following three cases.

\textbf{Case I.} 
Given $P\in[L,L\eta)$, we consider a partition $\calp := \{[L,L\eta), [L\eta,U]\}$ by letting $M_1 = L\eta$ and $M_2 = U$, and aim to show that $\ota$ is $(\eta,\gamma)$-competitive over $[L,L\eta)$ and $[L\eta,U]$.
In this case, we have $\Phi_P = L\eta = \phi(0) = \phi(1)$.
Therefore, the subset $[L,L\eta)$ belongs to Case I of the sufficient condition. Since $M_1/L \le \eta$, \ota is $\eta$-competitive over $[L,L\eta)$.
For subset $[L\eta,U]$, we have ${M_2 = U > \phi(1)}$. Based on the sufficient condition in Case III of Theorem~\ref{thm:ota-sufficient-condition} and knowing that $M_2/\int_0^1 \Phi_P du = U/(L\eta) = \theta/\eta = \gamma$, \ota is $\gamma$-competitive over $[L\eta,U]$.
Thus, \ota is $(\eta,\gamma)$-competitive over $\calp$. 

\textbf{Case II.} 
Given $P\in[L\eta,L\gamma)$, we consider a partition $\calp := \{[L,\Phi_P), [\Phi_P,P], (P,U]\}$ by letting $M_1 = \Phi_P$, $M_2 = P$ and $M_3 = U$.
Note that $P\ge \Phi_P$ since ${P = \eta P/\eta = \la \gamma P /\eta + (1-\la)P/\eta \ge \Phi_P}$ based on the design of $\eta$ and $\gamma$ in Equation~\eqref{eq:eta-gamma-design-1-max}.
We aim to show that \ota is $(\gamma,\eta,\gamma)$-competitive over $\calp$.
In this case, $\phi(0) = \phi(1) = \Phi_P = \la L\gamma + (1-\la)P/\eta$. 

For subset $[L,\Phi_P)$, we have $M_1 \le \phi(0)$ and we consider the sufficient condition in Case I. Since
\begin{align}
\frac{M_1}{L} = \frac{\Phi_P}{L} = \lambda\gamma + \frac{1-\lambda}{L\eta}P \le \lambda\gamma + (1-\lambda)\frac{\gamma}{\eta} \le \gamma,  
\end{align}
\ota is $\gamma$-competitive over $[L,\Phi_P)$.

For subset $[\Phi_P,P]$, if $P\in(L\eta,L\gamma)$, we have $M_2 = P > \phi(1) = \Phi_P$ and consider the sufficient condition in Case III. Since
\begin{align}
\frac{M_2}{\int_0^1 \Phi_P du} = \frac{P}{\Phi_P} = \frac{P}{\lambda L \gamma + \frac{1-\lambda}{\eta} P} \le \frac{1}{\lambda + \frac{1-\lambda}{\eta}} \le \eta, 
\end{align}
\ota is $\eta$-competitive over $[\Phi_P,P]$. If $P = L\eta$, we have $M_2 = \phi(0)$. Based on Case I of the sufficient condition and $M_2/L = \eta$, \ota is also $\eta$-competitive over $[\Phi_P,P]$. 

For subset $(P,U]$, we consider the sufficient condition of Case III, and we have
\begin{align}
\frac{M_3}{\int_0^1 \Phi_P du} = \frac{U}{\Phi_P} = \frac{U}{\lambda L \gamma + \frac{1-\lambda}{\eta} P}\le \frac{U}{\lambda L \gamma + (1-\lambda) L} = \frac{\theta}{\eta} = \gamma,
\end{align}
where we apply $\eta = \la \gamma + 1-\la$ in~\eqref{eq:eta-gamma-design-1-max}.
Therefore, \ota is $\gamma$-competitive over $(P,U]$.

Thus, \ota is $(\gamma,\eta,\gamma)$-competitive over $\calp$.

\textbf{Case III.} 
Given $P\in[L\gamma,U]$, 
we consider a partition $\calp := \{[L,L\gamma), [L\gamma,U]\}$ by letting $M_1 = L\gamma$, and aim to show \ota is $(\gamma,\eta)$-competitive over $\calp$.
For subsets $[L,L\gamma)$ and $[L\gamma,U]$, we have $M_1/L = \gamma$ and $U/\Phi_P = U/(L\gamma) =\eta$, respectively. Based on the sufficient condition in Case I and Case III, \ota is $(\gamma,\eta)$-competitive over $\calp$.

In above three cases, given any $P$, we have shown that there exists a partition and \ota is $\eta$-competitive for the instance subset that contains $\Omega_P$ and $\gamma$-competitive for the other subsets. Based on Lemma~\ref{lem:connection}, $\ota$ is $\eta$-consistent and $\gamma$-robust. 

\subsection{Proof of Theorem~\ref{thm:one-way-trading}}
\label{app:proof-one-way-trading}

We prove the competitiveness of \ota with the threshold function in~\eqref{eq:threshold-otp-P} for one-way trading based on the sufficient condition in Theorem~\ref{thm:ota-sufficient-condition}. Consider the following three cases.

\textbf{Case I.} Given $P\in[L,M)$, we consider a partition $\calp:=\{[L,M),[M,U]\}$, and aim to show that \ota is $(\eta,\gamma)$-competitive over $[L,M)$ and $[M,U]$. Let $\phi_1(w), w\in[0,\beta)$ and $\phi_2(w), w\in[\beta,1]$ denote the two pieces of the threshold functions given in~\eqref{eq:threshold-otp-P1}. 
Both $[L,M)$ and $[M,U]$ belong to Case II of the sufficient condition in Theorem~\ref{thm:ota-sufficient-condition}.

For subset $[L,M)$, $\phi_1(w) = L + (L\eta - L)\exp(\eta w)$ has no flat segment, i.e., $\beta_0' = \beta_0 = 0$, and its increasing segment $\varphi_1$ is the solution of the differential equation
\begin{align}
\begin{cases}
\varphi_1(w) = \eta \left[\int_{0}^{w}\varphi_1(u) du + (1-w)L\right],  w\in[0,\beta),\\
\varphi_1(\beta) = M,
\end{cases}
\end{align}
which satisfies the sufficient condition in~\eqref{eq:sufficient-condition-DE} if $M = L + (L\eta- L)\exp(\eta \beta)$. 

For subset $[M,U]$, $\phi_2(w) = L + (U-L)\exp(\gamma(w-1))$ also has no flat segment, i.e., $\beta_1 = \beta_1' = \beta$, and the increasing segment $\varphi_2$ is the solution of 
\begin{align}
\begin{cases}
\varphi_2(w) = \gamma\left[\int_{0}^{\beta}\phi(u)du + \int_{\beta}^w\varphi_2(u)du + (1-w)L\right],  w\in[\beta,1],\\
\varphi_2(1) = U,
\end{cases}
\end{align}
which satisfies the sufficient condition in~\eqref{eq:sufficient-condition-DE} if $M\gamma/\eta = L + (U-L)\exp(\gamma(\beta-1))$.

Since $M$ and $\beta$ are the solution of equation~\eqref{eq:otp-threshold-breakpoints1}, both $\phi_1$ and $\phi_2$ satisfy the sufficient condition in Case II. 
Thus, \ota is $(\eta,\gamma)$-competitive over $[L,M)$ and $[M,U]$

\textbf{Case II.} Given $P\in[M,U)$, we consider a partition $\calp:=\{[L,M_1),[M_1,P],(P,U]\}$, where $M_2 = P$, $M_3 = U$, and $M_1\in[M,P)$ is to be determined. 
We aim to show that \ota with $\phi$ in~\eqref{eq:threshold-otp-P2} is $(\gamma,\eta,\gamma)$-competitive over $\calp$. 
Let $\phi_1(w), w\in[0,\beta_1)$, $\phi_2(w), w\in[\beta_1,\beta_2]$, and $\phi_3(w), w\in(\beta_2,1]$ denote the threshold pieces corresponding to the three subsets.
Based on the threshold function~\eqref{eq:threshold-otp-P2} and $\beta_1$ and $\beta_2$ determined in equation~\eqref{eq:otp-threshold-breakpoints2}, we have $\phi(0) = \min\{M_1,L\gamma\}$, and $\phi(1) = P$ if $P\gamma/\eta \ge U$ and $\phi(1) = U$ if $P\gamma/\eta < U$.

For subset $[L,M_1)$, we consider the following two sub-cases based on the value of $\max\{M_1/L,\gamma\}$. 

\textbf{Case II(a).} If $M_1 \le L\gamma$, based on the first equation in~\eqref{eq:otp-threshold-breakpoints2}, $\beta_1 = 0$ and hence $\phi_1$ is absorbed. In this case, $M_1=\phi(0)$ and thus $[L,M_1)$ belongs to Case I of the sufficient condition. Since $M_1/L \le \gamma$, \ota is $\gamma$-competitive over $[L,M_1)$.

\textbf{Case II(b).} If $M_1 > L\gamma$, we have $\phi(0)< M_1 \le \phi(1)$ and hence $[L,M_1)$ belongs to Case II of the sufficient condition.
In this case, $\phi_1(w) = L + (\gamma L - L)\exp(\gamma w), w\in[0,\beta_1)$ has no flat segment, and the increasing segment $\varphi_1$ is the solution of 
\begin{align}
\begin{cases}
\varphi_1(w) = \gamma \left[\int_{0}^{w}\varphi_1(u) du + (1-w)L\right],  w\in[0,\beta_1),\\
\varphi_1(\beta_1) = M_1,
\end{cases}
\end{align}
which satisfies the sufficient condition in~\eqref{eq:sufficient-condition-DE} if $M_1 = L + (\gamma L - L)\exp(\gamma \beta_1)$.

Summarizing {Case II(a)} and {Case II(b)}, \ota is $\gamma$-competitive over $[L,M_1)$ if the first equation in~\eqref{eq:otp-threshold-breakpoints2} holds.

For subset $[M_1,P]$, since $\phi(0) < P \le \phi(1)$, $[M_1,P]$ belongs to Case II of the sufficient condition. 
$\phi_2(w) = M_1, w\in[\beta_1,\beta_1')$ is a flat segment and the sufficient condition in~~\eqref{eq:sufficient-condition-DE} holds when $w = \beta_1'$ if the length of this segment ensures $\phi_2(\beta_1') = M_1 = \eta [\int_{0}^{\beta_1}\phi(u)du + (\beta_1' - \beta_1)M_{1} + (1-\beta_1')L]$, which is the second equation in~\eqref{eq:otp-threshold-breakpoints2}.

The increasing segment $\varphi_2(w), w\in[\beta_1',\beta_2]$ is the solution of 
\begin{align}
\begin{cases}
\varphi_2(w) = \eta\left[\int_{0}^{\beta_1}\phi(u)du + (\beta_1' - \beta_1)M_{1} + \int_{\beta_1'}^w \varphi_2(u)du + (1-w)L \right], w\in[\beta_1',\beta_2],\\
\varphi_2(\beta_2) = P,
\end{cases}
\end{align}
which satisfies the sufficient condition in~\eqref{eq:sufficient-condition-DE} if $P = L + (M_1 - L)\exp(\eta(\beta_2 - \beta_1'))$. Thus, \ota is $\eta$-competitive over $[M_1,P]$ if the second and third equations in~\eqref{eq:otp-threshold-breakpoints2} hold.

For subset $(P,U]$, we have the following two sub-cases based on the value of $\min\{P\gamma/\eta,U\}$.

\textbf{Case II(c).} If $P\gamma/\eta < U$, we have $\phi(0)<U\le \phi(1) = U$ and thus $(P,U]$ belongs to Case II of the sufficient condition. $\phi_3(w) = L + (U-L)\exp(\gamma(w-1)),w\in(\beta_2,1]$ has no flat segment and its increasing segment $\varphi_3$ is the solution of 
\begin{align}
\begin{cases}
\varphi_3(w)  = \gamma \left[\int_{0}^{\beta_2}\phi(u)du + \int_{\beta_2}^w\varphi_3(u)du + (1-w)L \right], w\in(\beta_2,1],\\
\varphi_3(1) = U,
\end{cases}
\end{align}
which satisfies the sufficient condition~\eqref{eq:sufficient-condition-DE} if $P\gamma/\eta = L + (U-L)\exp(\gamma(\beta_2 - 1))$.

\textbf{Case II(d).} If $P\gamma/\eta \ge U$, we have $\beta_2 = 1$ based on the fourth equation in~\eqref{eq:otp-threshold-breakpoints2}. Thus, $\phi(1)= P< U$ and $(P,U]$ belongs to Case III of the sufficient condition. Since $U/\int_0^1\phi(u)du = U\eta/P \le \gamma$, \ota is $\gamma$-competitive over $(P,U]$.

Based on Case II(c) and Case II(d), \ota is $\gamma$-competitive over $(P,U]$ if the forth equation in~\eqref{eq:otp-threshold-breakpoints2} holds.

In summary, since $M_1$, $\beta_1$, $\beta_1'$ and $\beta_2$ are the solution of equation~\eqref{eq:otp-threshold-breakpoints2}, \ota with the threshold function~\eqref{eq:threshold-otp-P2} is $(\gamma,\eta,\gamma)$-competitive over $[L,M_1),[M_1,P],(P,U]$ .

\textbf{Case III.} Given $P = U$, we consider a partition $\calp:=\{[L,U), [U]\}$. Based on equation~\eqref{eq:otp-threshold-breakpoints2} with $P=U$, we have $\beta_2 = \beta_1' = 1$ and $M_1 = U$. Thus, $\phi_3$ and the increasing segment of $\phi_2$ are absorbed. Since $\phi(0) < U \le \phi(1)$, both $[L,U)$ and $[U]$ belong to Case II of the sufficient condition. 

For subset $[L,U)$, $\phi_1(w) = L+(\gamma L - L)\exp(\gamma w), w\in[0,\beta_1)$ has no flat segment and its increasing segment $\varphi_1$ is the solution of
\begin{align}
\begin{cases}
\varphi_1(w) = \gamma \left[\int_{0}^{w}\varphi_1(u) du + (1-w)L\right], w\in[0,\beta_1),\\
\varphi_1(\beta_1) = U,
\end{cases}
\end{align}
which satisfies the sufficient condition~\eqref{eq:sufficient-condition-DE} if $M_1 = U = L + (\gamma L - L)\exp(\gamma \beta_1)$.

For subset $[U]$, $\phi_2(w) = U, w\in[\beta_1,1]$ satisfies the sufficient condition in~\eqref{eq:sufficient-condition-DE} when $w = \beta_1' = 1$ if $\varphi_2(1) = U = \eta [\int_{0}^{\beta_1}\phi(u)du  + (1-\beta_1)U]$.
This equation holds if the first equation in~\eqref{eq:otp-threshold-breakpoints2} holds with $M_1 = U$, and $\eta = {\theta}/\left[\frac{\theta}{\gamma}+(\theta - 1)(1-\frac{1}{\gamma}\ln\frac{\theta- 1}{\gamma  - 1})\right]$, which holds based on equation~\eqref{eq:robust-consiste-otp}.

Since $\beta_1$ is the solution of the first equation in~\eqref{eq:otp-threshold-breakpoints2} with $M_1 = U$, \ota is $(\gamma,\eta)$-competitive over $[L,U)$ and $[U]$. 

With the competitiveness results in above three cases, \ota with the threshold function~\eqref{eq:threshold-otp-P} is $\eta$-consistent and $\gamma$-robust based on Lemma~\ref{lem:connection}.

\subsection{Proof of Theorem~\ref{thm:lower-bound-1-max}}
\label{app:proof-lower-bound-1-max}

Let $g(p):[L,U]\to\{0,1\}$ denote a conversion function of a deterministic online algorithm for 1-max-search, where $g(p) = 1$ (or $g(p) = 0$) represents converting $1$ (or $0$) dollar under the instance $\cali_p$ before the compulsory conversion in the last step.
Based on the same arguments as those for the conversion function of one-way trading, the conversion function of 1-max-search satisfies that (i) $g(p)$ is non-decreasing in $[L,U]$ and (ii) $g(U) = 1$.

Let $\cali_{\hat{\Phi}}$ denote the first instance, under which an online algorithm for 1-max-search converts $1$ dollar, where $\hat{\Phi} = \inf_{\{p\in[L,U]:g(p)=1\}} p$ is defined as the conversion price.
We claim $\hat{\Phi}$ of any $\gamma$-robust online algorithm is upper bounded by $\gamma L$. 
This claim can be proved by contradiction. 
Suppose the conversion price of a $\gamma$-robust algorithm is $\gamma L+\varepsilon, \varepsilon >0$. 
The profit ratio of the offline optimal and online algorithm under the instance $\cali_{\gamma L+\varepsilon/2}$ is $\opt(\cali_{\gamma L+\varepsilon/2})/\alg(\cali_{\gamma L+\varepsilon/2}) = (\gamma L+\varepsilon/2)/L > \gamma$, which contradicts with the $\gamma$-robustness of this algorithm.

Given a prediction $P = U$, to ensure $\eta$-consistency, any $\gamma$-robust online algorithm must have $\eta \ge \opt(\cali_{U})/\alg(\cali_{U}) = U/\hat{\Phi} \ge U/(\gamma L) = \theta/\gamma$, where the second inequality is due to the constraint $\hat{\Phi} \le \gamma L$ from $\gamma$-robustness. 

Based Theorem~\ref{thm:1-max-search}, \ota with the reservation price~\eqref{eq:threshold-1-max} achieves the robustness-consistency trade-off $\eta = \theta/\gamma$, which matches the lower bound, and thus is Pareto-optimal for 1-max-search.


\section{Detailed Experimental Setup}\label{subsec:app-exp-setup} 
 
 We use historical Bitcoin (BTC) prices in USD of 5 years from October 2015 through December 2020, with exchange rate information collected every 5 minutes.  Our dataset uses publicly available BTC exchange rates gathered from the Gemini cryptocurrency exchange.  BTC has gone through dramatic price highs and lows over the years, with a minimum exchange rate of \$353 and maximum exchange rate of \$29,305 by the end of 2020.  
 
 In the experiments, each instance captures a trading period of one week and assumes one unit of BTC is available to be traded.  Notably, BTC is traded 24/7, so each one week trading period is composed of $2016 = 12\times24\times7$ five-minute exchange rates.  Over the course of 5 years, there are 250 instances of 7 days.  In order to facilitate additional rounds in online learning experiments, \otpalgalf learns over 583 instances of 7 days that each overlap by 3 days.  If an instance is the period 1/01/2016 to 1/07/2016, the next overlapping instance is 1/04/2016 to 1/10/2016.

To generate a simple prediction $P$ of a one week instance, we use the observed maximum exchange rate of the previous week. With prediction error $\epsilon = |OPT - P|$, we also test the effect of varying prediction quality by adjusting $\epsilon$ offline with a multiplicative error level between 0 and 1.0, where 0 error level indicates perfect predictions and 1.0 level indicates unadjusted predictions. To evaluate the performance in worst-case settings, we also introduce a \textit{crash probability} $q$, where the exchange rate of BTC in the last timeslot of the one-week trading period is equal to the lower bound $L$ with probability $q$.  In fact, BTC experienced a drop of over \$19,000 in a single week of May 2021 following news of Tesla and financial institutions in China no longer accepting BTC as payment.

We report the empirical profit ratio, which is the profit of the optimal offline algorithm over the profit of an online algorithm.  This is the counterpart of the theoretical competitive ratio in the empirical setting. 

\end{appendices}

%% file: main.bbl
\begin{thebibliography}{10}

\bibitem{angelopoulos2019online}
S.~Angelopoulos, C.~D{\"u}rr, S.~Jin, S.~Kamali, and M.~Renault.
\newblock Online computation with untrusted advice.
\newblock In {\em Proc. of Innovations in Theoretical Computer Science
  Conference (ITCS)}, 2020.

\bibitem{antoniadis2020online}
A.~Antoniadis, C.~Coester, M.~Elias, A.~Polak, and B.~Simon.
\newblock Online metric algorithms with untrusted predictions.
\newblock In {\em International Conference on Machine Learning}, pages
  345--355. PMLR, 2020.

\bibitem{antoniadis2020secretary}
A.~Antoniadis, T.~Gouleakis, P.~Kleer, and P.~Kolev.
\newblock Secretary and online matching problems with machine learned advice.
\newblock In {\em 34th Conference on Neural Information Processing Systems},
  2020.

\bibitem{bamas2020primal}
E.~Bamas, A.~Maggiori, and O.~Svensson.
\newblock The primal-dual method for learning augmented algorithms.
\newblock {\em Advances in Neural Information Processing Systems}, 33, 2020.

\bibitem{blum2006online}
A.~Blum, T.~Sandholm, and M.~Zinkevich.
\newblock Online algorithms for market clearing.
\newblock {\em Journal of the ACM (JACM)}, 53(5):845--879, 2006.

\bibitem{el1998competitive}
R.~El-Yaniv.
\newblock Competitive solutions for online financial problems.
\newblock {\em ACM Computing Surveys (CSUR)}, 30(1):28--69, 1998.

\bibitem{el2001optimal}
R.~El-Yaniv, A.~Fiat, R.~M. Karp, and G.~Turpin.
\newblock Optimal search and one-way trading online algorithms.
\newblock {\em Algorithmica}, 30(1):101--139, 2001.

\bibitem{emek2011online}
Y.~Emek, P.~Fraigniaud, A.~Korman, and A.~Ros{\'e}n.
\newblock Online computation with advice.
\newblock {\em Theoretical Computer Science}, 412(24):2642--2656, 2011.

\bibitem{li2014online}
B.~Li and S.~C. Hoi.
\newblock Online portfolio selection: A survey.
\newblock {\em ACM Computing Surveys (CSUR)}, 46(3):1--36, 2014.

\bibitem{lorenz2009optimal}
J.~Lorenz, K.~Panagiotou, and A.~Steger.
\newblock Optimal algorithms for k-search with application in option pricing.
\newblock {\em Algorithmica}, 55(2):311--328, 2009.

\bibitem{lykouris2018competitive}
T.~Lykouris and S.~Vassilvtiskii.
\newblock Competitive caching with machine learned advice.
\newblock In {\em International Conference on Machine Learning}, pages
  3296--3305. PMLR, 2018.

\bibitem{maillard2010}
O.-A. Maillard and R.~Munos.
\newblock Online learning in adversarial lipschitz environments.
\newblock {\em Joint european conference on machine learning and knowledge
  discovery in databases}, pages 305--320, 2010.

\bibitem{Mitrinovic1991}
D.~S. Mitrinovic, J.~Pecaric, and A.~M. Fink.
\newblock {\em Inequalities involving functions and their integrals and
  derivatives}, volume~53.
\newblock Springer Science \& Business Media, 1991.

\bibitem{purohit2018improving}
M.~Purohit, Z.~Svitkina, and R.~Kumar.
\newblock Improving online algorithms via ml predictions.
\newblock {\em Advances in Neural Information Processing Systems},
  31:9661--9670, 2018.

\bibitem{sun2020competitive}
B.~Sun, A.~Zeynali, T.~Li, M.~Hajiesmaili, A.~Wierman, and D.~H. Tsang.
\newblock Competitive algorithms for the online multiple knapsack problem with
  application to electric vehicle charging.
\newblock {\em Proceedings of the ACM on Measurement and Analysis of Computing
  Systems}, 4(3):1--32, 2020.

\bibitem{wei2020optimal}
A.~Wei and F.~Zhang.
\newblock Optimal robustness-consistency trade-offs for learning-augmented
  online algorithms.
\newblock {\em Advances in Neural Information Processing Systems}, 33, 2020.

\bibitem{zhang2017optimal}
Z.~Zhang, Z.~Li, and C.~Wu.
\newblock Optimal posted prices for online cloud resource allocation.
\newblock {\em Proceedings of the ACM on Measurement and Analysis of Computing
  Systems}, 1(1):1--26, 2017.

\bibitem{zhou2008budget}
Y.~Zhou, D.~Chakrabarty, and R.~Lukose.
\newblock Budget constrained bidding in keyword auctions and online knapsack
  problems.
\newblock In {\em International Workshop on Internet and Network Economics},
  pages 566--576. Springer, 2008.

\end{thebibliography}
